%% file: main_cmame.tex
\documentclass[preprint, 11pt]{elsarticle}

\usepackage{amsxtra}
\usepackage{amsthm}
\usepackage{amssymb}
\usepackage{bm}
\usepackage{amsfonts}
\usepackage{algorithm,algcompatible}
\usepackage{graphicx}    
\usepackage{rotating}
\usepackage{color}
\usepackage{epsfig}
\usepackage{subfigure}   
\usepackage{verbatim}
\usepackage{natbib}   
\usepackage{float}
\usepackage[printonlyused]{acronym} 
\usepackage{natbib}

\input{color}

\begin{document}

\begin{frontmatter}

\title{A generalized likelihood-weighted optimal sampling algorithm for rare-event probability quantification}

\author{Xianliang Gong}
\author{Yulin Pan\corref{cor1}}
\ead{yulinpan@umich.edu}
\address{Department of Naval Architecture and Marine Engineering, University of Michigan, \\ 48109, MI, USA}
\cortext[cor1]{Corresponding author}

\begin{abstract}
In this work, we introduce a new acquisition function for sequential sampling to efficiently
quantify rare-event statistics of an input-to-response (ItR) system with given input probability
and expensive function evaluations. Our acquisition is a generalization of the likelihood-weighted (LW) acquisition \cite{sapsis2020output, sapsis2022optimal}, that was initially designed for the same purpose and then extended to many other applications. The improvement in our acquisition comes from the generalized form with two additional parameters, by varying which one can target and address two weaknesses of the original LW acquisition: (1) that the input space associated with rare-event responses is not sufficiently stressed in sampling; (2) that the surrogate model (generated from samples) may have significant deviation from the true ItR function, especially for cases with complex ItR function and limited number of samples. In addition, we develop a critical procedure in Monte-Carlo discrete optimization of the acquisition function, which achieves orders of magnitude acceleration compared to existing approaches for such type of problems. The superior performance of our new acquisition to the original LW acquisition is demonstrated in a number of test cases, including some cases that were designed to show the effectiveness of the original LW acquisition. We finally apply our method to an engineering example to quantify the rare-event roll-motion statistics of a ship in a random sea.  
\end{abstract}
\end{frontmatter}
\section{Introduction}
Rare events are generally abnormal system responses to some inputs that can occur in many physical and societal systems, often associated with catastrophic consequences. Typical examples include tsunamis, extreme precipitations, ship capsizes, and pandemic spikes. The quantification of rare-event statistics in system response is therefore of vital importance for the assessment and improvement of the system reliability \cite{farazmand2019extreme,ghil2011extreme,rahmstorf2011increase, pickering2022discovering, gong2023adaptive}.

In many applications, the system of interest can be characterized by an input-to-response (ItR) function $f(\mathbf{x})$ with known input probability $p_{\mathbf{x}}(\mathbf{x})$. The function $f(\mathbf{x})$ is usually expensive to evaluate through numerical simulations or physical experiments, restricting the number of samples that can be placed for function evaluations. In order to reduce the required number of samples in quantifying rare-event statistics, methods using importance sampling \cite{feng2023dense, zhao2016accelerated}, control variate \cite{yang2023adaptive} and large deviation theory \cite{tong2021extreme, tong2022optimization, dematteis2019extreme} have been developed and extensively studied. However, these methods usually target only a single metric, e.g., probability of response above a given threshold, thus lacking a general view of the rare-event statistics. In addition, they often deal with cases where rare events of interest occur in an isolated region of the input space, which is not necessarily true for a complex function $f(\mathbf{x})$. 

In this paper, we will instead focus on a different type of approach relying on a surrogate (or meta) model of the function $f(\mathbf{x})$, which in principle overcomes the issues mentioned above. The learning of the surrogate can be achieved by, say, Gaussian process regression (GPR), but needs to be conducted with limited data (i.e., samples with function evaluations). A typical method involved here is active learning (or sequential sampling) which sequentially selects the next sample that is most informative to the quantity of interest. Pertaining to the calculation of rare-event statistics, i.e., the tail part of the response PDF $p_f(f)$, a successful method developed in the past few years is the likelihood-weighted (LW) sampling \cite{sapsis2020output, sapsis2022optimal, blanchard2020output}, in which the next optimal sample $\mathbf{x}^*$, given existing dataset $\mathcal{D}$, is selected as the one that maximizes an acquisition function 
\begin{equation}
    acq_{LW}(\mathbf{x})  = {\rm{var}}(f(\mathbf{x})|\mathcal{D}) w(\mathbf{x}),
\label{a-LW}
\end{equation}
where the LW factor $w(\mathbf{x}) = p_{\mathbf{x}}(\mathbf{x})/p_{\hat{f}}(\hat{f}(\mathbf{x}))$ is the ratio of input probability to predicted output probability with $\hat{f} \equiv \mathbb{E}(f|\mathcal{D})$ the surrogate model. The idea of using LW factor in acquisition has later been extended to many different applications beyond its original purpose of rare-event statistics evaluation, including rare-event forecasting \cite{pickering2022discovering, rudy2023output}, Bayesian optimization \cite{blanchard2021bayesian}, robot path planning \cite{blanchard2020informative}, multi-arm bandit \cite{yang2022output}, and has been adapted to multi-fidelity context \cite{gong2022multi}.

\begin{figure}
    \centering
    \includegraphics[width=0.8\linewidth]{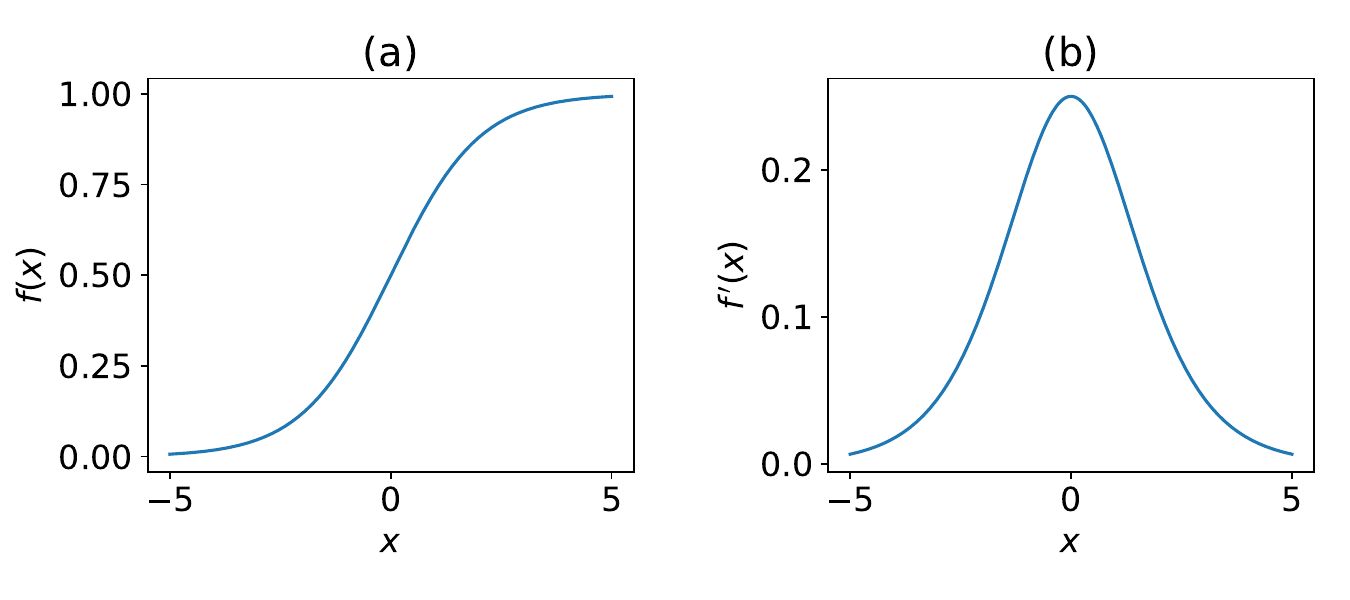}
    \caption{Plots of (a) a logistic function $f(x) = 1 / (1 + e^{-x})$ and (b) its derivative.}
    \label{fig:logistic}
\end{figure}

The effectiveness of the LW acquisition can be understood from the LW factor $w(\mathbf{x})$ in \eqref{a-LW}. It is argued in \cite{sapsis2020output} that due to $w(\mathbf{x})$, the next sample is chosen in favor of $\mathbf{x}$ with larger input probability $p_{\mathbf{x}}(\mathbf{x})$ (thus contributing more to $p_f(f)$) and smaller predicted response probability $p_{\hat{f}}(\hat{f}(\mathbf{x}))$ (thus associated with rare events). Such samples are more likely to contribute more to the rare-event (or tail) portion of the response PDF $p_f(f)$. While this interpretation is plausible, \eqref{a-LW} is clearly not an optimal sampling criterion. To see this, let us consider any one-dimensional monotonic function $f(x)$, say a logistic function $f(x) = 1 / (1 + e^{-x})$ (figure \ref{fig:logistic}(a)) and assume no difference between surrogate $\hat{f}(x)$ and ground-truth $f(x)$. The critical weighting factor is now reduced simply to $w(x)=f'(x)$, which peaks at $x=0$ (figure \ref{fig:logistic}(b)). It is clear not only that the input leading to large (and usually rare) responses is not stressed, but also that $w(x)$ has nothing to do with the input and response probability, violating the above claims made in \cite{sapsis2020output}. The failure of \eqref{a-LW} in the above example lies in the fact that the simple ratio in $w(\mathbf{x})$ is not necessarily optimal in sampling to resolve the tail of $p_f(f)$. Indeed, while larger $p_{\mathbf{x}}(\mathbf{x})$ and smaller $p_{\hat{f}}(\hat{f}(\mathbf{x}))$ helps, there is no guarantee that the optimal form is their direct ratio. In addition, another more severe issue regarding \eqref{a-LW} is that $\hat{f}$ may have a significant deviation from $f$, which makes small $p_{\hat{f}}(\hat{f}(\mathbf{x}))$ a poor estimation of the rarity of response. This is especially the case when function $f(\mathbf{x})$ is complex, given the limited number of samples that can be afforded. In such cases, the rare-event statistics provided by sampling through \eqref{a-LW} may become misleading since regions associated with small $p_f(f(\mathbf{x}))$ may never be explored (if the corresponding $p_{\hat{f}}(\hat{f}(\mathbf{x}))$ is large). 

In this work, we propose a generalized LW acquisition targeting the two limitations mentioned above. Our new acquisition takes a generalized form of \eqref{a-LW} with two additional parameters, by varying which one can achieve $(i)$ optimal deployment of $p(\mathbf{x})$ and $p_{\hat{f}}(\hat{f}(\mathbf{x}))$ in the LW factor, and $(ii)$ a much more effective guidance of sample exploration when $\hat{f}$ is very different from $f$. The generalized LW acquisition also shares the theoretical property of \eqref{a-LW} in terms of its derivation from the first principle. In addition, we point out an acceleration in Monte-Carlo discrete optimization regarding the acquisition, achieving orders of magnitude speedup compared to existing algorithms used in \cite{sapsis2020output, blanchard2020output, blanchard2021bayesian}. The superior performance of our new acquisition is consistently demonstrated in a number of test cases, including a stochastic oscillator \cite{blanchard2020informative, sapsis2022optimal}, a pandemic spike model \cite{pickering2022discovering} and cases with arbitrary complex functions $f(\mathbf{x})$ generated by kernels. We finally show the application of the new acquisition in an engineering example of quantifying the rare-event roll-motion statistics of a ship in a random sea.

The python code for the algorithm, named GPextreme, is available on Github\footnote{https://github.com/umbrellagong/GPextreme}.  

\section{Problem Statement}
We consider an ItR system described by a response function $f(\mathbf{x}):\mathbb{R}^{d} \to \mathbb{R}$ with input $\mathbf{x}$ a $d$-dimensional decision variable over a compact set and response an observable of the system. The input probability $p_\mathbf{x}(\mathbf{x})$ is assumed to be known, and our quantity of interest is the probability density function (PDF) of the response $p_f(f)$ with an emphasis on the tail part (to be precisely defined later in \eqref{error_1}). While $p_f(f)$ can be directly evaluated via standard Monte-Carlo method, an accurate resolution of its tail part is extremely expensive considering the expensiveness of system evaluations and the rareness of samples contributing to the PDF tail.

To reduce the computational cost, we make use of surrogate modeling with $f$ approximated by a learned surrogate (regressor) $\hat{f}$, achieved through Gaussian process regression (GPR) in this work. Assume we have a dataset $\mathcal{D}=\{\mathbf{X}, \mathbf{y}\}$ consisting of $n$ inputs $\mathbf{X} = \{\mathbf{x}^{i} \in \mathbb{R}^d\}_{i=1}^{n}$ and the corresponding outputs $\mathbf{y} = \{f(\mathbf{x}^i)\in \mathbb{R}\}_{i=1}^{n}$. In GPR, the underlining function is inferred as a posterior Gaussian process 
    $f(\mathbf{x})|\mathcal{D} \sim \mathcal{GP}\big(\mathbb{E}(f(\mathbf{x})|\mathcal{D}), {\rm{cov}}(f(\mathbf{x}), f(\mathbf{x}')|\mathcal{D})\big)$
with the mean as the surrogate, i.e., $\hat{f} \equiv \mathbb{E}(f|\mathcal{D})$ (see \cite{rasmussen2003gaussian} or \ref{app:gpr} for detailed formulae). With $\hat{f}$ available, the response PDF can be estimated as $p_{\hat{f}}(f)$ via evaluating $\hat{f}$ on a large number of samples (say standard Monte-Carlo samples), and our objective is to minimize the estimation error defined as (see \cite{mohamad2018sequential})
\begin{equation} 
    \epsilon = \int_{\Omega} \Big|\log p_{\hat{f}}(f) - \log p_{f}(f) \Big| df,
\label{error_1}
\end{equation}
where the integral is computed over a finite domain $\Omega = \mathrm{supp}(p_f) \cup \, \mathrm{supp}(p_{\hat{f}})$. We note that the $\log$ function in \eqref{error_1} acts on the ratio $p_{\hat{f}}(f)/p_{f}(f)$, which is amplified when $p_{f}(f)$ is small, i.e., the tail part of the PDF is reached.

Our goal is to construct $\hat{f}$ with a limited number of samples, i.e., to choose the most informative samples to learn $\hat{f}$ which facilitates the convergence of \eqref{error_1}. To achieve this goal, we use the idea of active learning (or sequential sampling) where the next sample is selected optimally based on the existing data $\mathcal{D}$. Specifically in the general form, the next-best sample is sequentially determined based on the optimization of an acquisition function:
\begin{equation}
    \mathbf{x}^* = \mathrm{argmax}_{\tilde{\mathbf{x}}} \; acq(\tilde{\mathbf{x}}; f(\mathbf{x})|\mathcal{D}),
\label{opt}
\end{equation}
with the overall algorithm detailed in Algorithm \textbf{1}. We will discuss the form of the acquisition function in \eqref{opt} in this paper, as the core of the algorithm. 

\begin{algorithm}
    \caption{Sequential active learning for rare-event statistics}
  \begin{algorithmic}
    \REQUIRE Number of initial samples $n_{init}$ and sequential samples $n_{seq}$
    \item[\textbf{Input:}] Initial dataset $\mathcal{D}=\{{\mathbf{x}^i}, f(\mathbf{x}^i) \}_{i=1}^{n_{init}}$ 
    \STATE \textbf{Initialization} $i = n_{init}$
    \WHILE{$i < n_{seq} + n_{init}$}
      \State Train GPR $f(\mathbf{x})|\mathcal{D}$
      \STATE Solve \eqref{opt} to find the next-best sample location $\mathbf{x}^{i+1}$
      \STATE Implement simulation/experiment to get $ f(\mathbf{x}^{i+1})$
      \STATE Update the dataset  $\mathcal{D} = \mathcal{D} \cup \{\mathbf{x}^{i+1}, f(\mathbf{x}^{i+1})\}$
      \STATE $i = i + 1$
    \ENDWHILE
\item[\textbf{Output:}] Compute the response PDF $p_{\hat{f}}$ based on the surrogate model $\hat{f}$
  \end{algorithmic}
\label{al}
\end{algorithm}

\section{Methodology Regarding Acquisition}
\subsection{Likelihood-weighted acquisition}
In using \eqref{error_1} as the error metric to guide the next sample, an issue comes up since the true function $f(\mathbf{x})$, thus $p_f(f)$, is unknown. To overcome this issue, \cite{mohamad2018sequential} proposed an effective proxy to \eqref{error_1} as 
\begin{equation} 
    \epsilon_{L}(\tilde{\mathbf{x}})  = \int |\log p_{f^{+}|\mathcal{D}, \hat{f}(\tilde{\mathbf{x}})}(f_{}) - \log p_{f^{-}|\mathcal{D}, \hat{f}(\tilde{\mathbf{x}})}(f_{})| \mathrm{d} f_{},
\label{e_L}
\end{equation}
\sloppy where $\tilde{\mathbf{x}}$ is the hypothetical location of the next sample, $p_{f^{\pm}|\mathcal{D}, \hat{f}(\tilde{\mathbf{x}})}(f)$ are PDF bounds generated by upper and lower bounds (say two standard deviations away from the mean) of GPR $f|\mathcal{D}, \hat{f}(\tilde{\mathbf{x}}) \sim \mathcal{GP}\big(\mathbb{E}(f(\mathbf{x})|\mathcal{D}, \hat{f}(\tilde{\mathbf{x}})), {\rm{cov}}(f(\mathbf{x}), f(\mathbf{x}')|\mathcal{D}, \hat{f}(\tilde{\mathbf{x}}))\big)$. However, in minimizing \eqref{e_L} with the next sample $\tilde{\mathbf{x}}$, one needs to compute $p_{f^{\pm}|\mathcal{D}, \hat{f}(\tilde{\mathbf{x}})}(f)$ for many $\tilde{\mathbf{x}}$, which is an expensive operation. To avoid this high computational cost, \cite{sapsis2020output} and  \cite{sapsis2022optimal} further constructed an upper bound of $\epsilon_L$ (up to a constant factor)
\begin{align}
    \epsilon_{LW}(\tilde{\mathbf{x}}) & = \int {\rm{var}}(f(\mathbf{x})|\mathcal{D}, \hat{f}(\tilde{\mathbf{x}})) \frac{p_{\mathbf{x}}(\mathbf{x})}{p_{\hat{f}}(\hat{f}(\mathbf{x}))}\mathrm{d} \mathbf{x},
\label{MSE-LW}
\end{align}
where $w(\mathbf{x}) \equiv p_{\mathbf{x}}(\mathbf{x}) / p_{\hat{f}}(\hat{f}(\mathbf{x}))$ is the LW factor with its significance reviewed in \S 1. Unlike the situation in \eqref{e_L}, the predicted response PDF $p_{\hat{f}}(f)$ only needs to be evaluated once in \eqref{MSE-LW}. 

The derivation of \eqref{MSE-LW} assumes that the surrogate $\hat{f}$ is sufficiently close to the true $f$. Under this assumption, an asymptotic form of $\epsilon_{L}$ can be first derived as (see \ref{app:approx_L} for a summary of the derivation following \cite{mohamad2018sequential} and \cite{sapsis2022optimal} but with clarifications of some critical procedures)
\begin{equation}
    \epsilon_{L} (\tilde{\mathbf{x}}) 
     \leq   \int \mathrm{std}(\mathbf{x}|\mathcal{D}, \hat{f}(\tilde{\mathbf{x}})) \frac{p_{\mathbf{x}}(\mathbf{x}) |p'_{\hat{f}}(\hat{f}(\mathbf{x}))| }{p^2_{\hat{f}}(\hat{f}(\mathbf{x}))}  \mathrm{d} \mathbf{x},
\label{approx_L}
\end{equation}
where $\mathrm{std}$ denotes standard deviation. With Cauchy-Schwarz inequality, $\epsilon_{L}$ can be further formulated as (see (3.4) in \cite{sapsis2022optimal})
\begin{equation}
    \epsilon_{L}(\tilde{\mathbf{x}}) \leq [\int \frac{p_{\mathbf{x}}(\mathbf{x}) p_{\hat{f}}'^2(\hat{f}(\mathbf{x}))}{p_{\hat{f}}^3(\hat{f}(\mathbf{x}))}\mathrm{d} \mathbf{x}]^{1/2} [\int {\rm{var}}(f(\mathbf{x})|\mathcal{D},\hat{f}(\tilde{\mathbf{x}})) \frac{p_{\mathbf{x}}(\mathbf{x})}{p_{\hat{f}}(\hat{f}(\mathbf{x}))}\mathrm{d} \mathbf{x}]^{1/2}.
\label{CSI}
\end{equation}
In \eqref{CSI}, the first term reduces to a constant and the second term squared leads to \eqref{MSE-LW}.

The next sample can be selected by minimizing \eqref{MSE-LW}, i.e., to construct the acquisition function in \eqref{opt} as $-\epsilon_{LW}(\tilde{\mathbf{x}})$. Alternatively, a more inexpensive but almost equally effective way (as tested in \cite{blanchard2020output}) is to choose the next sample at $\mathbf{x}$ which maximizes the integrand of \eqref{MSE-LW} without using the hypothetical sample $\tilde{\mathbf{x}}$, since getting sample there is supposed to contribute most significantly in reducing \eqref{MSE-LW}. Under the latter approach, we solve an optimization problem $\mathbf{x}^*=\mathrm{argmax}_\mathbf{x} \; acq_{LW}(\mathbf{x})$, with the acquisition function constructed as
\begin{equation}
    acq_{LW}(\mathbf{x}) = {\rm{var}}(f(\mathbf{x})|\mathcal{D}) \frac{p_{\mathbf{x}}(\mathbf{x})}{p_{\hat{f}}(\hat{f}(\mathbf{x})},
\label{US-LW}
\end{equation}
which is exactly \eqref{a-LW} in \S 1. \eqref{US-LW} can also be considered as the standard uncertainty sampling acquisition with a weighting factor inspired by \eqref{MSE-LW}. We further note that another advantage of \eqref{US-LW} over \eqref{MSE-LW} is that if neural networks are used to construct the surrogate model, \eqref{MSE-LW} involves excessive computational cost since ${\rm{var}}(f(\mathbf{x})|\mathcal{D}, \hat{f}(\tilde{\mathbf{x}}))$ does not have an analytical formulation as in GPR and needs to be re-trained for each $\tilde{\mathbf{x}}$ \cite{pickering2022discovering, guth2023evaluation}. Given the simplicity and effectiveness of \eqref{US-LW}, we will establish most of our analysis based on \eqref{US-LW}, but will discuss the derivation leading to \eqref{MSE-LW} (that inspires \eqref{US-LW}) in the subsequent parts of the paper.

\subsection{Proposed generalization of the LW acquisition}
The LW acquisition \eqref{US-LW} outperforms the other existing acquisitions in rare-event statistics quantification in several cases presented in \cite{blanchard2020output, sapsis2020output, sapsis2022optimal}, and it has a theoretical foundation outlined in \S 3.1. However, the insufficiency of \eqref{US-LW} discussed in \S 1 (e.g., discussion regarding figure \ref{fig:logistic}) is also intuitively true. How can we reconcile these two views on the LW acquisition \eqref{US-LW}?

In fact, the two limitations discussed in \S 1 roots exactly from the derivation of \eqref{MSE-LW}. First, from \eqref{approx_L} the Cauchy-Schwarz inequality can be applied in many different ways, and \eqref{CSI} is not necessarily the unique form of the upper bound. To be more specific, the integrand of \eqref{approx_L} can be distributed into two factors in many different ways, resulting in the fact that the second term in \eqref{CSI} may yield arbitrary powers on $p_{\mathbf{x}}(\mathbf{x})$ and $p_{\hat{f}}(\hat{f}(\mathbf{x}))$. Indeed, from this derivation itself, any of these resultant forms can serve as an upper bound to \eqref{approx_L} and none of them is unique. This is consistent with our intuitive argument in \S 1 that the direct ratio between $p_{\mathbf{x}}$ and $p_{\hat{f}}$ is not necessarily the optimal. Second, the derivation leading to \eqref{approx_L} relies on the assumption that $\hat{f} \approx f$. As discussed in \S 1, this is not necessarily true especially for complex function $f$, considering limited number of samples that can be placed. In case that $\hat{f}$ misses a region of large (usually rare) responses of interest, this region may never get explored by using \eqref{US-LW} since the associated $p_{\hat{f}}$ is not small. 

To address the above two limitations, we propose a generalization of LW acquisition \eqref{US-LW}, in the form of
\begin{equation}
    acq_{GLW}(\mathbf{x}) = {\rm{var}}(f(\mathbf{x}|\mathcal{D})) \big(w_G(\mathbf{x}, t, 0) + w_G(\mathbf{x}, t, \alpha) + w_G(\mathbf{x}, t, -\alpha)\big) ,
\label{US-GLW}
\end{equation}
where
\begin{align}
    w_G(\mathbf{x}, t, \alpha)  & = p_\mathbf{x}(\mathbf{x}) / p_{\hat{f}_\alpha}(\hat{f}_\alpha(\mathbf{x}))^t, \\
    \hat{f}_\alpha(\mathbf{x}) & = \hat{f}(\mathbf{x}) + \alpha \, {\rm{std}}(f(\mathbf{x})|\mathcal{D}).
\label{w_G}
\end{align}
which contains two additional parameters $t$ and $\alpha$. Regarding the first limitation, the parameter $t$ controls the level of emphasis on small $p_{\hat{f}}$ in the LW factor, and provides flexibility in balancing the need to sample at large-$p_\mathbf{x}$ and small-$p_{\hat{f}}$ region. With $t=1$, the first term in \eqref{US-GLW} reduces to the original LW acquisition \eqref{US-LW}. For $t>1$ and $t<1$, \eqref{US-GLW} places respectively more and less emphasis on small $p_{\hat{f}}$, i.e., the rarity of predicted response. We note that larger value of $t$ (i.e., more emphasis on small $p_{\hat{f}}$) does not mean better performance, since the performance needs to be eventually judged by the error metric \eqref{error_1}. Regarding the second limitation, the second and third terms in \eqref{US-GLW} provide more exploration power for the acquisition function. If $\hat{f}$ misses some large responses at $\mathbf{x}$, the GPR at these $\mathbf{x}$ is certain to be associated with large variance, so that either $f_{\alpha}$ or $f_{-\alpha}$ captures the large responses and plays an active role in \eqref{US-GLW}.


While the inclusion of $t$ and $\alpha$ in \eqref{US-GLW} provides flexibility in addressing the limitations in \eqref{US-LW}, the optimal values of these parameters cannot be theoretically determined (at least from the theoretical framework reviewed in this paper) and must depend on specific features of the function $p_{\mathbf{x}}(\mathbf{x})$ and $f(\mathbf{x})$. Therefore, the optimal $t$ and $\alpha$ values can only be empirically obtained through numerical tests as we will discuss in \S 4.  

\subsection{Acceleration in optimization of the acquisition functions}

In solving the optimization problem regarding the acquisition \eqref{US-LW} (and thus the generalized form \eqref{US-GLW}), the Monte Carlo discrete optimization (MCDO) method has been considered as an effective approach, which is tested to be superior to gradient-based method due to the non-convexity of the acquisitions in many cases \cite{pickering2022discovering}. In MCDO method, a large number of candidate samples located at $\mathbf{X}_{mc}\in \mathbb{R}^{n_{mc}*d}$ (usually from space-filling L-H sampling) are created, with $n_{mc} \gg n$ (with $n$ the number of samples in $\mathcal{D}$), from which one selects the candidate that returns a maximum in the acquisition. Such procedure allows all $acq(\mathbf{X}_{mc})$ to be evaluated in one vector operation that saves much computational cost than other global or gradient-based optimization methods that rely on iterations. We also note that in optimization regarding \eqref{MSE-LW}, function evaluation on pre-selected Monte-Carlo samples $\mathbf{X}_{mc}$ is also needed in evaluation of the integral, as conducted in \cite{sapsis2020output, blanchard2020output, blanchard2021bayesian}. Therefore, the acceleration method we introduce below applies equally to the optimization problems regarding \eqref{MSE-LW}, \eqref{US-LW}, and \eqref{US-GLW}. 

In computing $acq(\mathbf{X}_{mc})$, say with \eqref{US-LW}, one needs to evaluate a new GPR with $\hat{f}(\mathbf{X}_{mc}) = \mathbb{E}(\mathbf{X}_{mc}|\mathcal{D})$ and $\mathrm{var}(\mathbf{X}_{mc}|\mathcal{D})$, with the former needed to calculate the function $p_{\hat{f}}(f)$. In obtaining these quantities, \cite{blanchard2020output, blanchard2021bayesian} have suggested to apply the recursive formula such that the new GPR can be built recursively with new data $\mathbf{x}_n$ leveraging previous GPR based on $\mathcal{D}_{n-1}$, instead of a brute-force retraining taking all $\mathcal{D}$. In doing so, the previous works argued that the computational cost can be much reduced compared to brute-force retraining. However, in the context of MCDO method, a careful analysis conducted in \ref{app:opt} shows that the retraining process (in particular the inverse of covariance on $\mathcal{D}$) only constitutes a very small portion of the total computational cost considering $n_{mc} \gg n$. Therefore, the bottleneck of the computation in fact comes from the prediction step, that is the generation of the covariance matrix and multiplication of matrices involving $n_{mc}$ rows/columns. In order to overcome this major part of the computational cost, we develop a matrix re-grouping technique (that is in analogy to the regrouping technique used in many adjoint methods) and apply the idea of memory-time tradeoff on top of the recursive formula. With details and test cases presented in \ref{app:opt}, we show that the original computational complexity $O(n_{mc}*n^2)$ (which holds with or without simply applying the recursive formula) can be reduced to $O(n_{mc}*n)$. This is a significant reduction considering $n \gtrsim O(100)$ in many applications.

\section{Results}
In this section, we test the performance of generalized LW acquisition $acq_{GLW}$ in \eqref{US-GLW} with variations of $\alpha$ and $t$, to show its advantage over $acq_{LW}$. The test cases are organized as follows: In \S4.1, we choose two models with simple response functions that were previously used for demonstrating the effectiveness of $acq_{LW}$ in \cite{blanchard2020output, sapsis2022optimal} and  \cite{pickering2022discovering}. We shall show that using $acq_{GLW}$ (especially with appropriate $t$) achieves additional significant benefits in reducing the error defined in \eqref{error_1}. In \S4.2, we use as response functions a large number $O(1000)$ of synthetic functions from realizations of Gaussian processes, with most functions complex with multi-modes. We demonstrate the advantage of $acq_{GLW}$ over $acq_{LW}$ especially with appropriate value of $\alpha$. In \S4.3, we consider the application of $acq_{GLW}$ to an engineering problem of evaluating the rare-event statistics of ship motion in a random sea.

\subsection{Two test cases in existing works}
\subsubsection{Stochastic oscillator}
\begin{figure}
    \centering
    \begin{minipage}[b]{0.46\linewidth}
    \includegraphics[width = \linewidth]{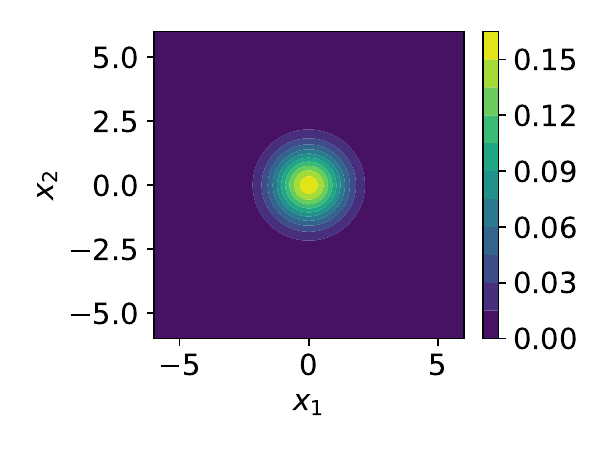}
    \centering{(a)}
    \end{minipage}
    \begin{minipage}[b]{0.46\linewidth}
    \includegraphics[width = \linewidth]{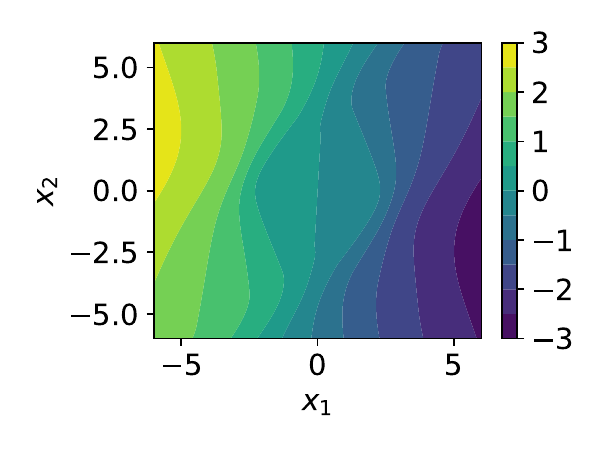}
    \centering{(b)}
    \end{minipage}
     \vskip.5\baselineskip
    \begin{minipage}[b]{0.6\linewidth}
    \includegraphics[width = \linewidth]{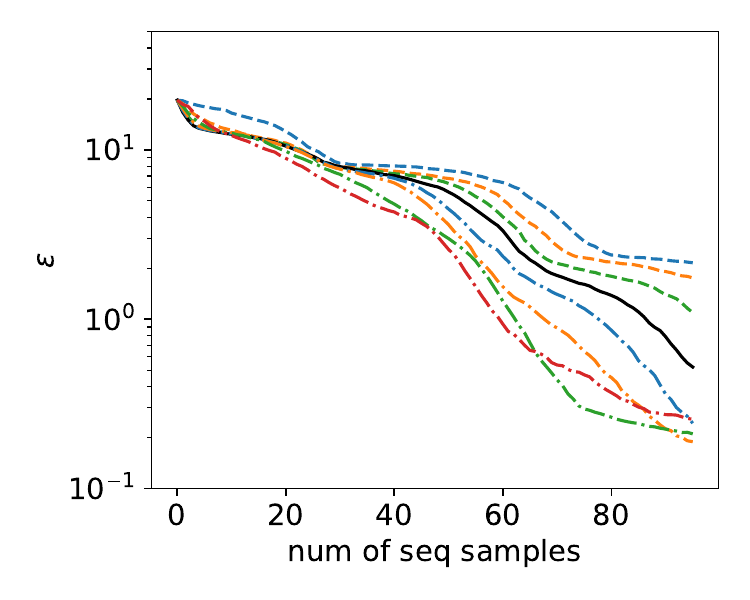}
    \centering{(c)}
    \end{minipage}
    \caption{(a) input probability distribution and (b) response function of the stochastic oscillator example. (c) error $\epsilon$ as functions of sample numbers with $\alpha=0$ and varying $t=$ 0.6 (\bluedashedline), 0.8(\orangedashedline), 0.9(\greendashedline), 1(\blackline), 1.1(\bluedashdotline), 1.2(\orangedashdotline), 1.4(\greendashdotline), 1.6(\reddashdotline).}
    \label{fig:oscillator}
\end{figure}

The first case we choose consists of a 2D response function constructed from the solution of a nonlinear oscillator under stochastic excitation, which is studied in \cite{blanchard2020output, sapsis2022optimal}. In particular, the oscillator equation is formulated as
\begin{equation} 
    \ddot{u}(t) + \delta \dot{u}(t) + F(u) = \xi(t),
\end{equation}
where $u(t)$ is the state variable, $F$ is a nonlinear restoring force defined as
\begin{align}
F(u) = \left\{
\begin{aligned}
    & \alpha u           &
    &  if \; 0 \leq |u| \leq u_1 \\
    & \alpha u_1         &
    &  if \; u_1 \leq |u| \leq u_2  \\
    & \alpha u_1 + \beta(u-u_2)^3 &
    &  if \; u_2 \leq|u|           \\
\end{aligned}
\right..
\end{align} 
The stochastic process $\xi(t)$, with a correlation function $\sigma_{\xi}^2 e^{-\tau^2/(2 l_{\xi}^2)}$, is approximated by a two-term Karhunen-Loeve expansion
\begin{equation} 
    \xi(t) = \sum_{i=1}^{2} x_i \lambda_i \phi_i(t),
\end{equation}
with $\lambda_i$ and $\phi_i(t)$ respectively the eigenvalue and eigenfunction of the correlation function, $\mathbf{x}\equiv (x_1, x_2)$ is a standard normal variable as the input to the system (see figure \ref{fig:oscillator}(a)), satisfying $p_{\mathbf{x}}(\mathbf{x}) = \mathcal{N}(\mathbf{0}, \mathrm{I}_2)$ with $\mathrm{I}_2$ being a $2 \times 2$ identity matrix. The values of the parameters are kept the same as those in the existing works\footnote{$\delta$=1.5, $\alpha$=1, $\beta$=0.1, $u_1$=0.5, $u_2$=1.5, $\sigma_{\xi}^2$=0.1, $l_\xi$=4.}.

\begin{figure}
    \centering
     \includegraphics[width=\linewidth]{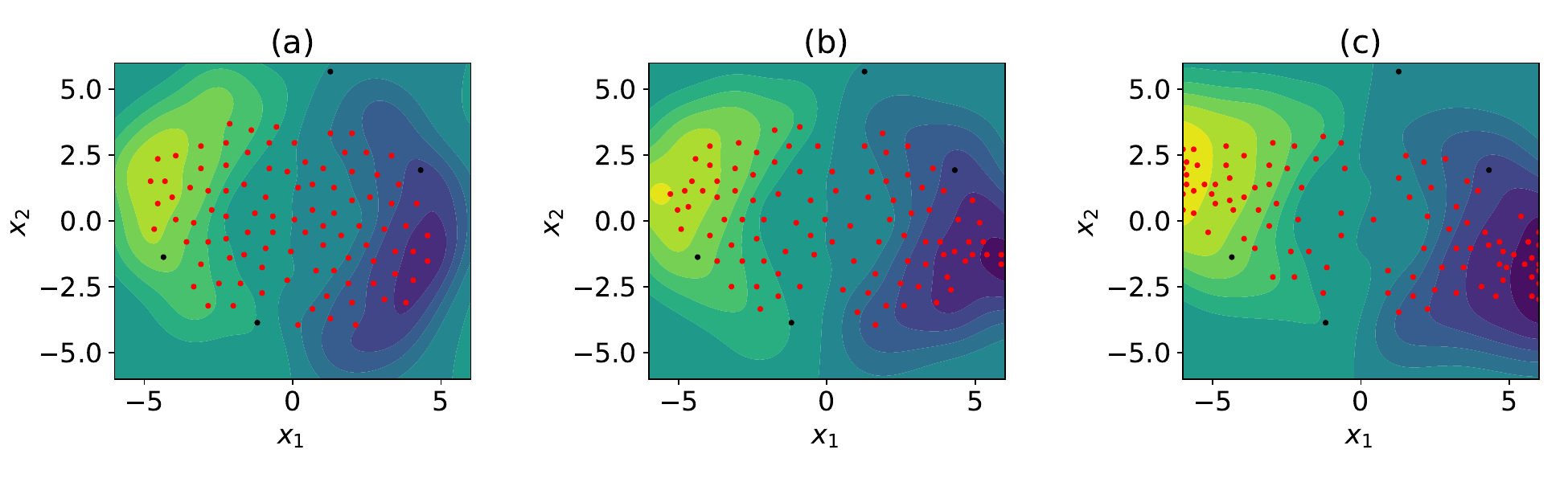}
    \caption{Predicted response functions and sequential sampling locations (\tikzcircle{1pt, red}) with $\alpha=0$ and (a) $t=$ 0.6, (b) $t=$ 1, (c) $t=$ 1.4 in the stochastic oscillator example, starting from the same initial samples (\tikzcircle{1pt, black}).}
    \label{fig:os_samples}
\end{figure}

The response of the system is considered as the mean value of $u(t;\mathbf{x})$ in the interval $[0,25]$:
\begin{equation} 
    f(\mathbf{x}) = \frac{1}{25} \int_{0}^{25}u(t;\mathbf{x}) \mathrm{d} t, 
\label{oscillator}
\end{equation}
with contour shown in figure \ref{fig:oscillator}(b).

In our computation, we use 4 initial samples followed by 96 sequential samples with the error metric $\epsilon$ in \eqref{error_1} calculated after each sample. Considering the randomness of initial samples, all results show below are average from 100 different initializations unless otherwise specified. Figure \ref{fig:oscillator}(c) shows $\epsilon$ as a function of sample number for different values of $t$ in $acq_{GLW}$, including the case of $t=1$ for which $acq_{GLW}$=$acq_{LW}$\footnote{We note that our result with $t=1$ is different from that in \cite{blanchard2020output}. This is because \cite{blanchard2020output}, for some reason, sets a floor value of $e^{-16}$ for $p_f(f)$ in their calculation, which is unnecessarily high for double precision. We instead set a floor value of $10^{-16}$ that is consistent with double precision.}. We see that the optimal performance of $acq_{GLW}$ is achieved for $t$ roughly in $[1.2, 1.6]$, where $\epsilon$ is about half an order of magnitude smaller than that with $t=1$ close to the end of sampling. The favorable performance with $t\in [1.2, 1.6] $ can be further understood from the sample locations shown in figure \ref{fig:os_samples}. As expected, when $t$ is increased from 0.6 to 1.4, more samples are allocated in the input space with extreme-value responses, leading to a smaller error $\epsilon$ characterizing the accuracy of the tail of the response PDF.

\subsubsection{Pandemic spike} 
\begin{figure}
    \centering
    \begin{minipage}[b]{0.46\linewidth}
    \includegraphics[width = \linewidth]{figure/oscillator/input_os.pdf}
    \centering{(a)}
    \end{minipage}
    \begin{minipage}[b]{0.46\linewidth}
    \includegraphics[width = \linewidth]{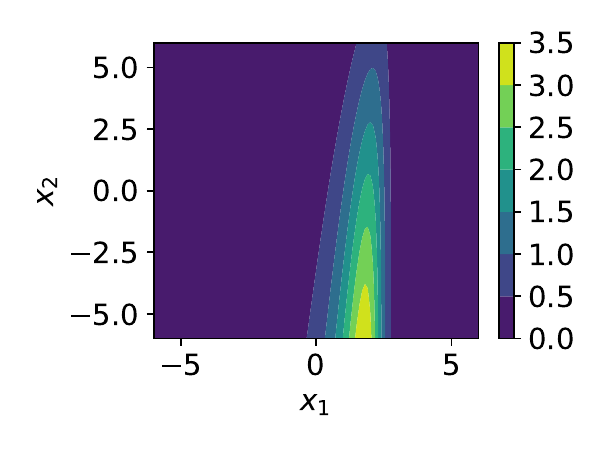}
    \centering{(b)}
    \end{minipage}
     \vskip.5\baselineskip
    \begin{minipage}[b]{0.6\linewidth}
    \includegraphics[width = \linewidth]{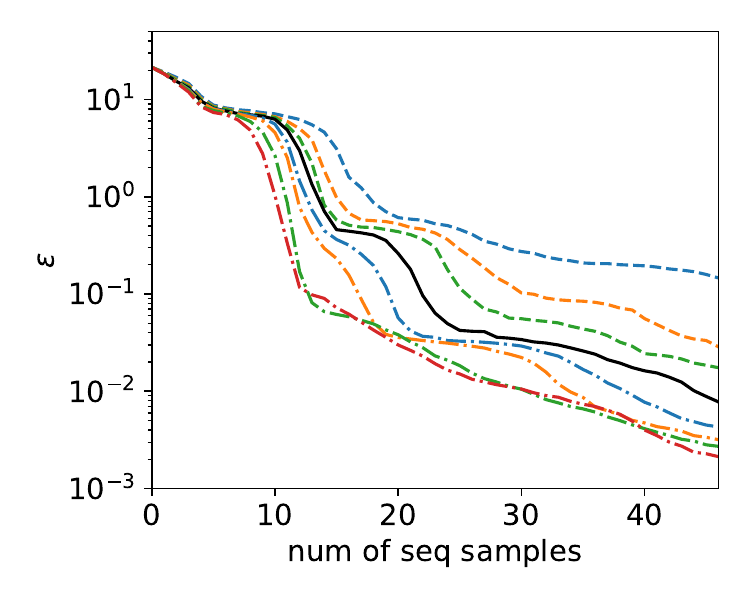}
    \centering{(c)}
    \end{minipage}
    \caption{(a) input probability distribution and (b) response function of the pandemic spike example. (c) error $\epsilon$ as functions of sample numbers with $\alpha=0$ and varying $t=$ 0.6 (\bluedashedline), 0.8(\orangedashedline), 0.9(\greendashedline), 1(\blackline), 1.1(\bluedashdotline), 1.2(\orangedashdotline), 1.4(\greendashdotline), 1.6(\reddashdotline).}
    \label{fig:sir}
\end{figure}

We consider another case used in \cite{pickering2022discovering}, where the response function is constructed from the evolution of infections in a pandemic. In particular, the evolution of infections is simulated by Susceptible, Infected, Recovered (SIR) model developed in \cite{kermack1927contribution} and \cite{anderson1979population}
\begin{align}
    \frac{\mathrm{d}S}{\mathrm{d}t} & = - \beta I S + \delta R
\nonumber \\
    \frac{\mathrm{d}I}{\mathrm{d}t} & = \beta I S - \gamma I 
\nonumber \\ 
    \frac{\mathrm{d}R}{\mathrm{d}t} & = \gamma I  - \delta R, 
\label{sir}
\end{align}
with $S(t)$, $I(t)$, and $R(t)$ respectively state variables representing the number of susceptible, infectious and recovered individuals. $\delta$, $\gamma$, and $\beta$ are immunity loss rate, recovery rate, and infection rate. The parameter $\beta$ is endowed with a two-term K-L expansion of the stochastic process: $\beta(t) = \beta_0(\sum_{i=1}^{2} x_i \lambda_i \phi_i(t) + \phi_0)$ with $\phi_0>0$ and $\phi_i(t)$, $\lambda_i$ determined from the correlation function $\sigma_{\beta}^2 e^{-\tau^2/(2 l_{\beta}^2)}$. We keep all parameter values and initial conditions to \eqref{sir} the same as in \cite{pickering2022discovering}\footnote{$\delta=0$, $\gamma=0.1$, $\beta_0=3*19^{-9}$, $\phi_0=2.55$, $\sigma^2_{\beta}=0.1$, $l_{\beta}=4$, $S(0)=10^8$, $I(0) =50$, $R(0)=0$.}. The input variable $\mathbf{x}\equiv (x_1, x_2)$ is a standard normal variable with $p_{\mathbf{x}}(\mathbf{x}) = \mathcal{N}(\mathbf{0}, \mathrm{I}_2)$ (see figure \ref{fig:sir}(a)). We are interested in, as the response of the system, the infections at $t=20$:
\begin{equation} 
    f(\mathbf{x}) = I(t=20; \mathbf{x}), 
\label{oscillator}
\end{equation}
with its contour shown in figure \ref{fig:sir}(b).

\begin{figure}
    \centering
     \includegraphics[width=\linewidth]{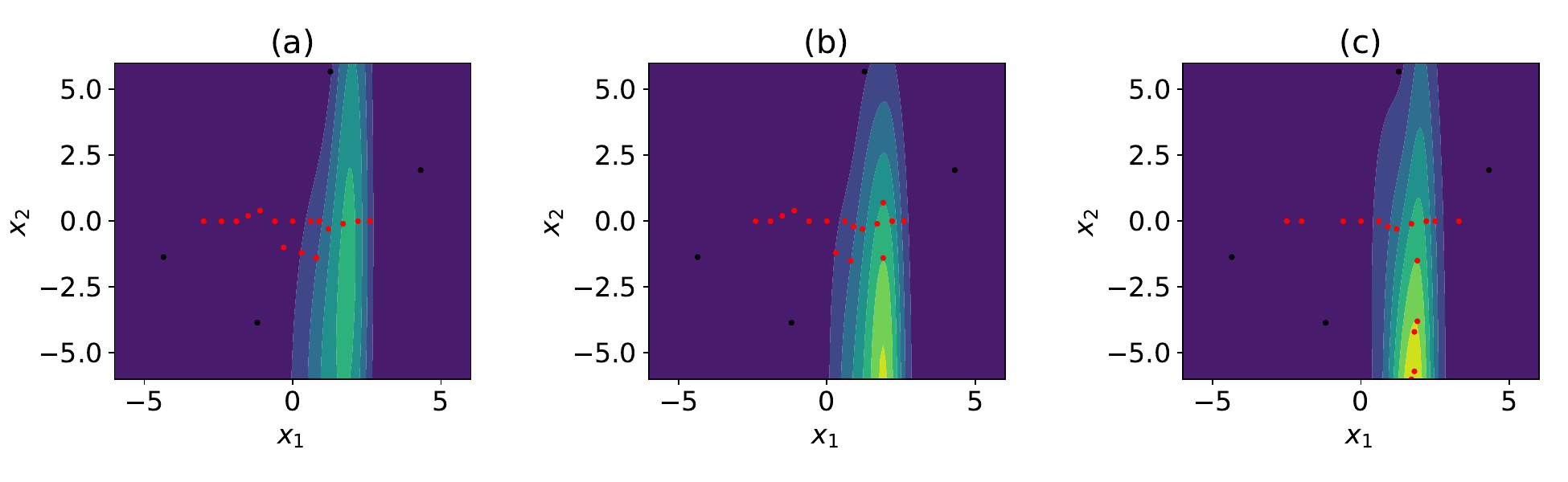}
    \caption{Predicted response functions and sequential sampling locations (\tikzcircle{1pt, red}) with $\alpha=0$ and (a) $t=$ 0.6, (b) $t=$ 1, (c) $t=$ 1.4 in the pandemic spike example, starting from the same initial samples (\tikzcircle{1pt, black}).}
    \label{fig:sir_samples}
\end{figure}

Our computation starts from 4 initial samples, followed by 46 sequential samples employing $acq_{GLW}$. The results with $\alpha=0$ and varying $t$ from $0.6$ to $1.6$ are plotted in figure \ref{fig:sir}(c) as a function of the number of sequential samples. We see a similar pattern as in \S4.1 where the optimal performance occurs with $t$ roughly in $[1.2, 1.6]$, for which the error $\epsilon$ at majority of sample numbers is about half an order of magnitude smaller than that obtained in the case with $t=1$. Furthermore, the sample locations for $t=$ 0.6, 1, and 1.4 plotted in figure \ref{fig:sir_samples} again show that the increase of $t$ pushes more samples toward rare-event regions in the input space.

We note that for the above two cases (and other cases with relatively simple response functions), the variation of $\alpha$ can also have an impact on the performance of sequential sampling. In particular, we have observed some cases with $\alpha>0$ that produce somewhat better results than those with $\alpha=0$. However, the mechanism associated with $\alpha$ is much more subtle than that with $t$ for these simple response functions, and we will not elaborate it in this paper. The impact of $\alpha$ on the sampling performance is most evident for complex (multi-modal) response functions, which we discuss in detail in \S4.2.

\subsection{Complex response functions generated by kernels}

\begin{figure}
    \centering
    \begin{minipage}[b]{\linewidth}
    \includegraphics[width = \linewidth]{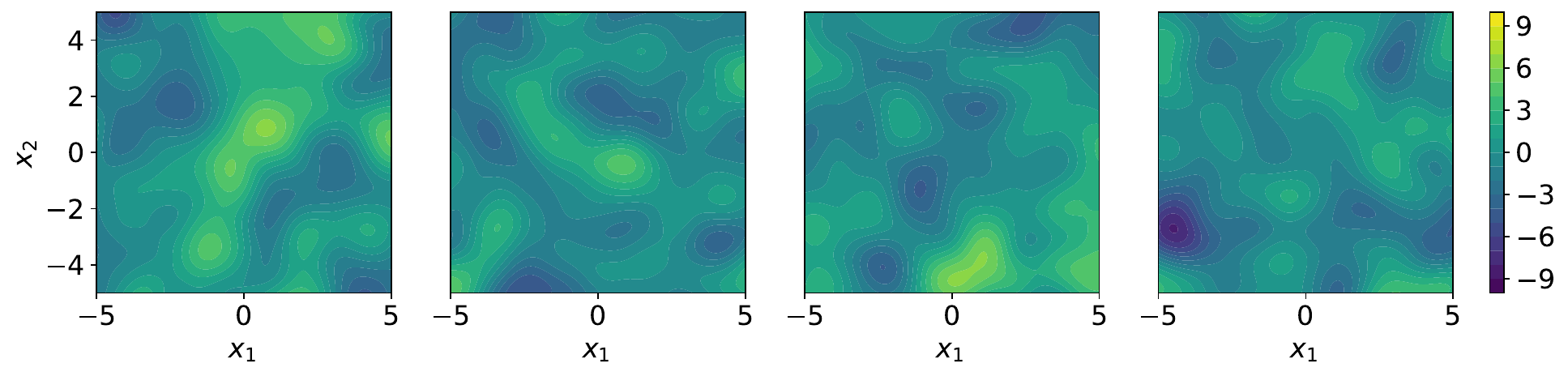}
    \centering{(a)}
    \end{minipage}
    \vskip.5\baselineskip
    \begin{minipage}[b]{\linewidth}
    \includegraphics[width = \linewidth]{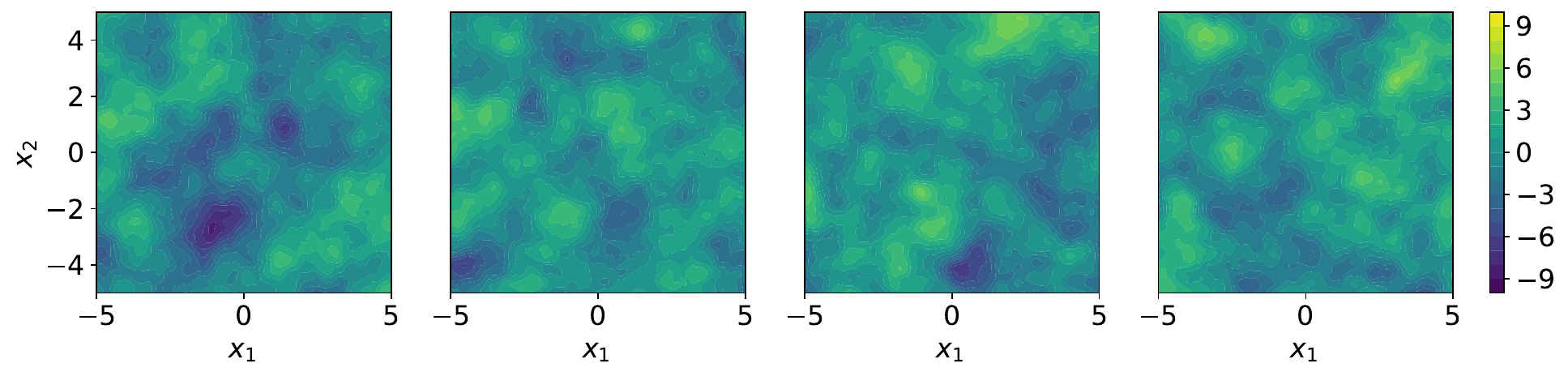}
    \centering{(b)}
    \end{minipage}
    \caption{Examples of two-dimensional (a) RBF and (b) Matern functions.} 
    \label{fig:2d_cases}
\end{figure}
In this section, we test the performance of $acq_{GLW}$ for a large number of arbitrarily-generated complex response functions. These functions are constructed as random realizations of Gaussian processes with RBF kernel and Matern kernel\footnote{The hyperparameters of kernels are set as $\tau=2$ and $\Lambda=\mathrm{I}_d$, and for Matern kernel the additional parameter $\nu$ is fixed as 1.5 (see \ref{app:gpr} for the definition of kernels and parameters).}, hereafter referred to as RBF and Matern functions for simplicity. Examples of such functions in the 2D case are shown in figure \ref{fig:2d_cases}, which illustrates the complex and multi-modal features of these functions (especially for Matern functions which exhibits more small-scale variations). In the tests presented below, we consider both 2D ($d=2$) and 3D ($d=3$) cases, with the input set as a standard normal $p_{\mathbf{x}}(\mathbf{x}) = \mathcal{N}(\mathbf{0}, \mathrm{I}_d)$. For each kernel and dimension, the results presented in terms of error $\epsilon$ are averaged over 200 function realizations of the random process and 20 different realizations of initial samples (for each function), i.e., over 4000 cases in total. Due to this massive average, the improved results from $acq_{GLW}$ presented below is statistically significant. For clarity of the presentation, we will show results for the RBF functions in the main text, and leave results for the Matern functions that lead to similar conclusions in \ref{app:Matern}. 


\subsubsection{Two-dimensional (2D) RBF functions}

\begin{figure}
    \centering
    \begin{minipage}[b]{0.48\linewidth}
    \includegraphics[width = \linewidth]{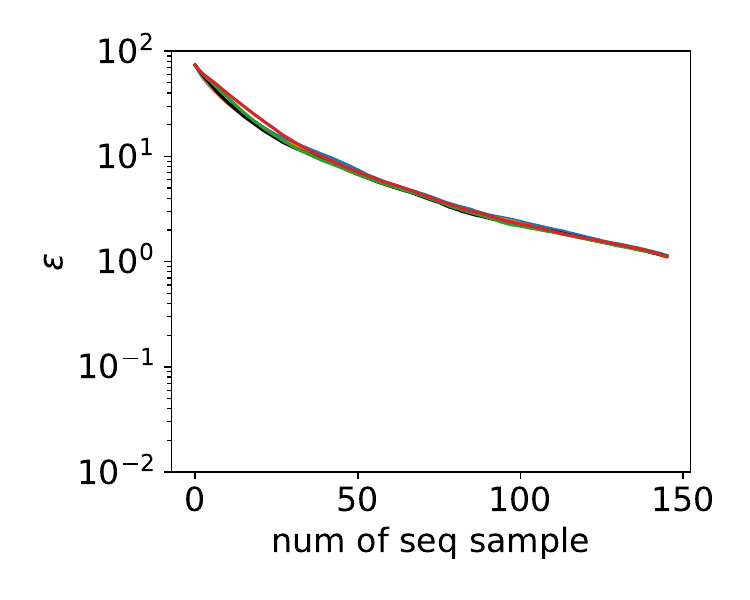}
    \centering{\quad (a)}
    \end{minipage}
    \begin{minipage}[b]{0.48\linewidth}
    \includegraphics[width = \linewidth]{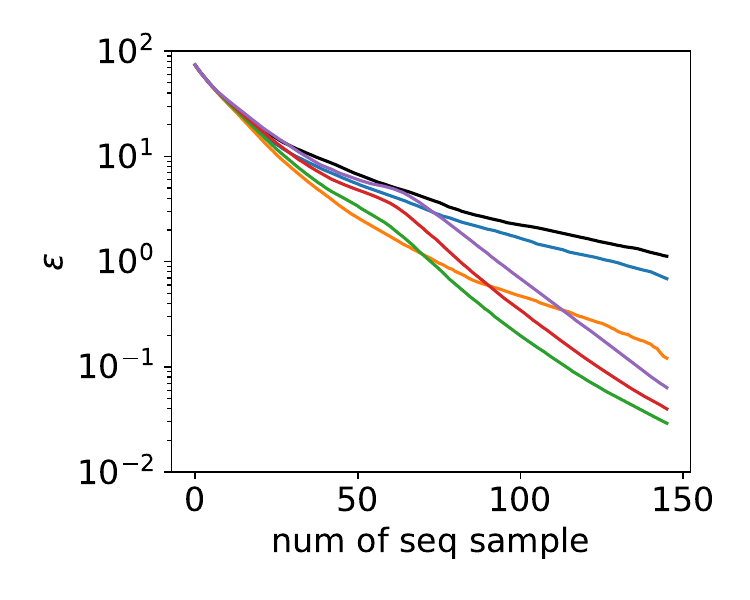}
    \centering{\quad (b)}
    \end{minipage}
    \begin{minipage}[b]{0.6\linewidth}
    \includegraphics[width = \linewidth]{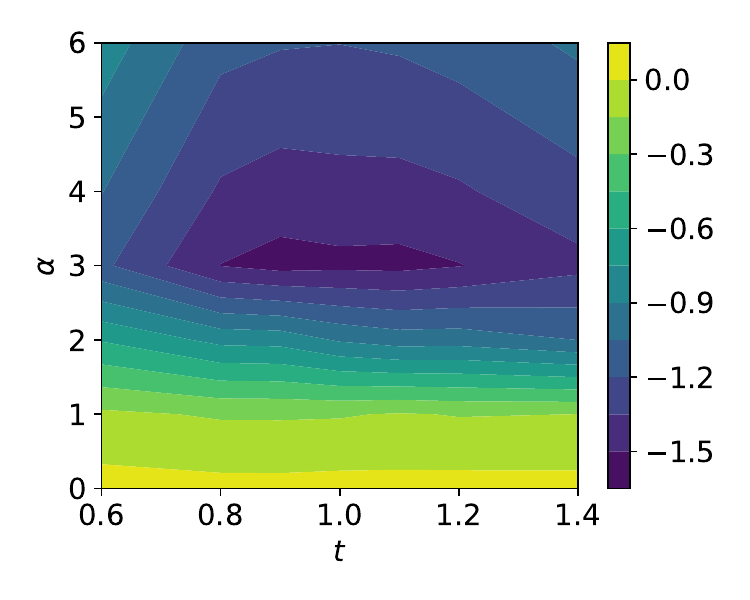}
    \centering{(c)}
    \end{minipage}
    \caption{Results for two-dimensional RBF functions. Error $\epsilon$ as function of number of samples for (a) $\alpha=0$ and varying $t=$ 0.6 (\blueline), 0.8(\orangeline), 1(\blackline), 1.2(\greenline), 1.4(\redline), (b) $t=1$ and varying $\alpha=$ 0(\blackline),  1 (\blueline), 2(\orangeline), 3(\greenline), 4(\redline), 6(\purpleline); (c) contour plot of $\log_{10} \epsilon$ at 146 sequential samples for varying $t$ and $\alpha$.
    }
    \label{fig:2d_RBF}
\end{figure}

\begin{figure}
    \centering
    \includegraphics[width=\linewidth]{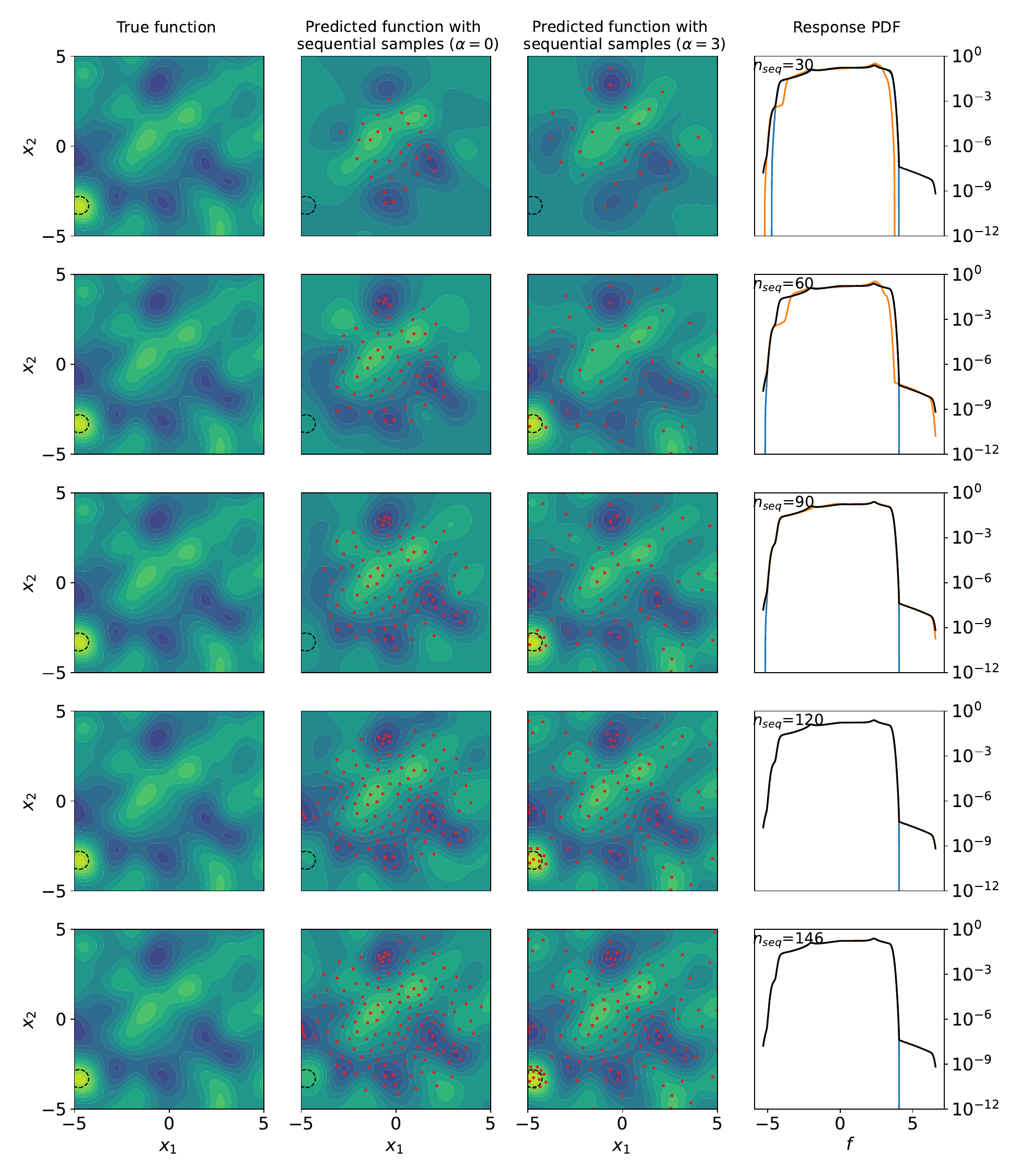}
    \caption{First column: true response RBF function as a reference; second column: sequential samples (\tikzcircle{2pt, Red}) with $\alpha=0$ on the predicted response function; third column: sequential samples (\tikzcircle{2pt, Red}) with $\alpha=3$ on the predicted response function; fourth column: predicted PDF $p_{\hat{f}}(f)$ with $\alpha=0$ (\blueline) and $\alpha=3$ (\orangeline) compared with the true PDF $p_f(f)$ (\blackline). The top-to-bottom rows correspond to situations with number of sequential samples $n_{seq} = [30, 60, 90, 120, 146]$. The black circles shown in columns 1-3 mark the rare-event region around $(-4.7, -3.3)$ that is missed by sequential samples with $\alpha=0$ but captured with $\alpha=3$. } 
    \label{fig:evolution_RBF}
\end{figure}
We first consider 2D RBF functions with examples plotted in figure \ref{fig:2d_cases}, showing much stronger variations (i.e., higher complexities) than the cases presented in \S 4.1. Figure \ref{fig:2d_RBF}(a) shows the error $\epsilon$ with increase of number of samples (4 initial samples followed by 146 sequential samples) for $t$ varying from 0.6 and 1.4 and fixed $\alpha=0$. Unlike the cases in \S 4.1, the variation of parameter $t$ almost does not affect the performance of sequential sampling using $acq_{GLW}$, with results indistinguishable for the selected range of $t$. On the other hand, variation of $\alpha$ leads to a much stronger impact on the error $\epsilon$, as shown in figure \ref{fig:2d_RBF}(b). One can see from the figure that $\alpha=3$ produces the best result in the tested range, with error at 146 sequential samples about two orders of magnitude smaller than that with $\alpha=0$ (i.e., the original $acq_{LW}$ acquisition). Figure \ref{fig:2d_RBF}(c) further shows a contour plot of the error $\epsilon$ at 146 sequential samples as a function of $t$ and $\alpha$. We see that $\alpha\approx 3$ and $t\approx 1$ is indeed close to the global optimal among all choices tested here. 

Considering the behavior observed in figure \ref{fig:2d_RBF}, it is clear that the improved performance associated with larger $\alpha$ comes from the increased exploration power of the acquisition that captures more rare-event regions in the input space. Such exploration is not achievable by the variation of $t$, at least in the tested range. To demonstrate this reasoning, we plot in figure \ref{fig:evolution_RBF} the evolution of sampling locations, predicted response functions, and predicted response PDFs from 30 to 146 sequential samples with $\alpha=0$ and $\alpha=3$ ($t=1$ fixed) for a typical RBF function. It is clear that with $\alpha=0$ the rare-event region near $\mathbf{x} = (-4.7, -3.3)$ (which happens to be the global maximum) is not captured, leading to a failure in resolving the right tail of the response PDF. More specifically, the missing of this important rare-event region is due to the fact that the predicted response $\hat{f}$ fails to capture the large response in this region with limited number of samples, together with the lack of exploration power with $\alpha=0$. In contrast, for $\alpha=3$, the region near $\mathbf{x} = (-4.7, -3.3)$ is identified within 60 sequential samples, leading to a much more accurate resolution of the right tail of the PDF. We encourage the readers to also take a look at figure \ref{fig:evolution_matern} for Matern response functions where such behavior is more evident due to the increased complexity of the function.

\subsubsection{Three-dimensional (3D) RBF functions}

\begin{figure}
    \centering
    \includegraphics[width=\linewidth]{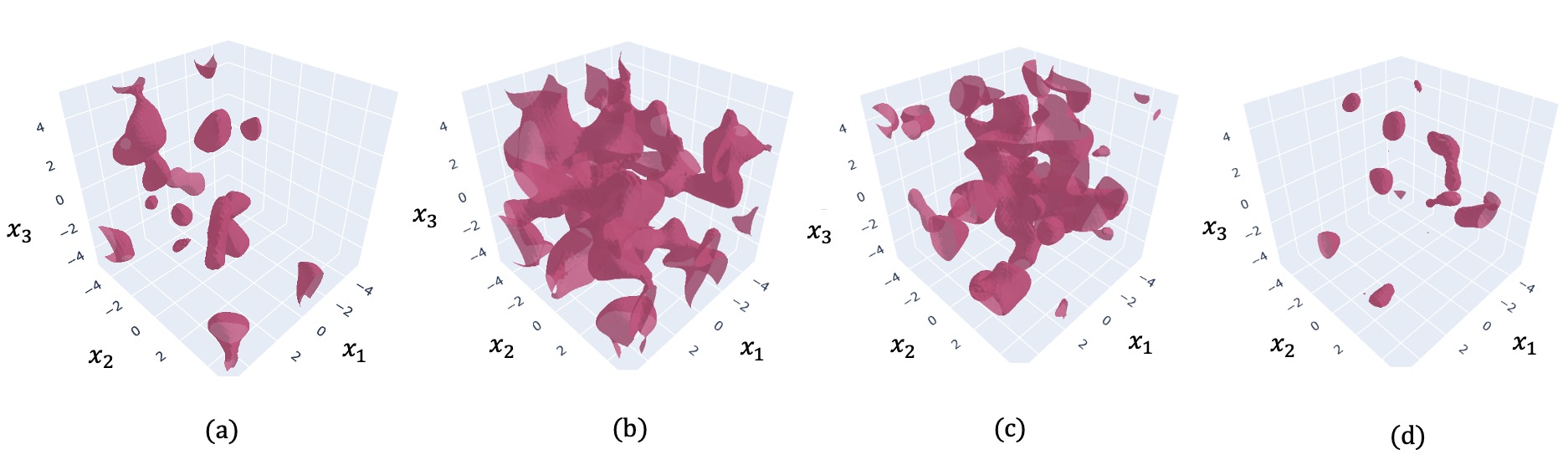}
    \caption{Level sets of a typical 3D RBF function. From (a) to (d), $\{\mathbf{x}: f(\mathbf{x}) = -4, -2, 2, 4\}$}
    \label{fig:3d_func}
\end{figure}

\begin{figure}
    \centering
    \begin{minipage}[b]{0.48\linewidth}
    \includegraphics[width = \linewidth]{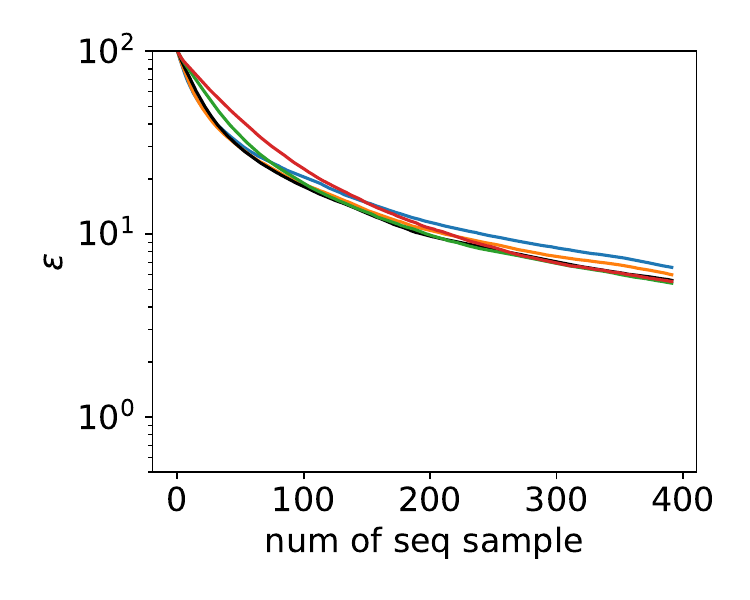}
    \centering{\quad \quad (a) }
    \end{minipage}
    \begin{minipage}[b]{0.48\linewidth}
    \includegraphics[width = \linewidth]{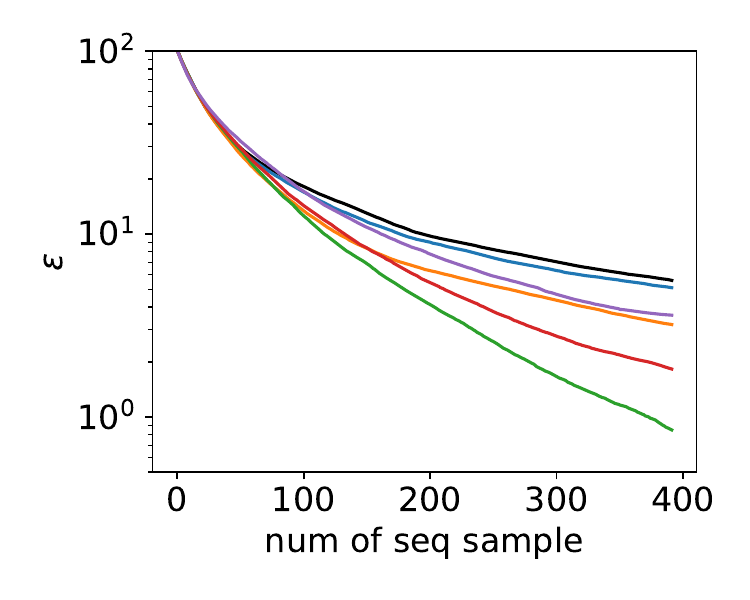}
    \centering{\quad \quad  (b) }
    \end{minipage}
    \begin{minipage}[b]{0.6\linewidth}
    \includegraphics[width = \linewidth]{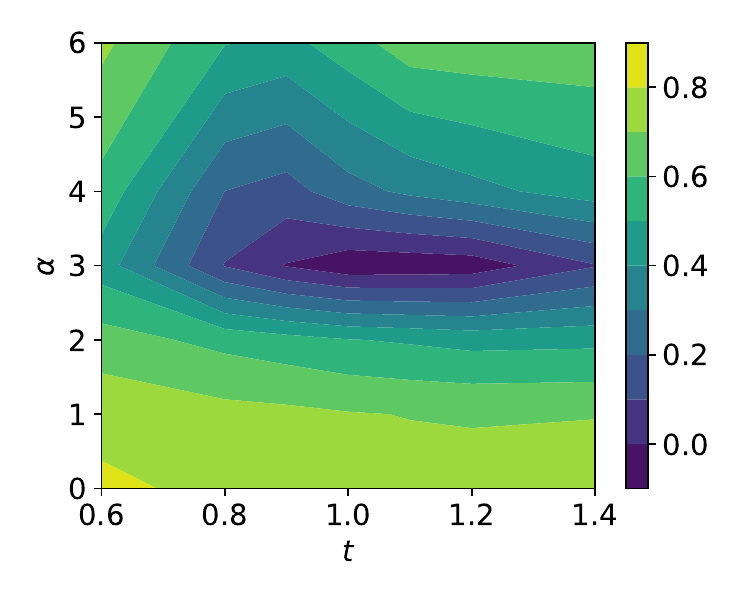}
    \centering{(c)}
    \end{minipage}
    \caption{Results for three-dimensional RBF functions. Error $\epsilon$ as function of number of samples for (a) $\alpha=0$ and varying $t=$ 0.6 (\blueline), 0.8(\orangeline), 1(\blackline), 1.2(\greenline), 1.4(\redline), (b) $t=1$ and varying $\alpha=$ 0(\blackline),  1 (\blueline), 2(\orangeline), 3(\greenline), 4(\redline), 6(\purpleline); (c) contour plot of $\log_{10} \epsilon$ at 392 sequential samples for varying $t$ and $\alpha$.
    }
    \label{fig:3d_RBF} 
\end{figure}

\begin{figure}
    \centering
    \begin{minipage}[b]{\linewidth}
    \includegraphics[trim={0 0 3cm 0},clip, width=\linewidth]{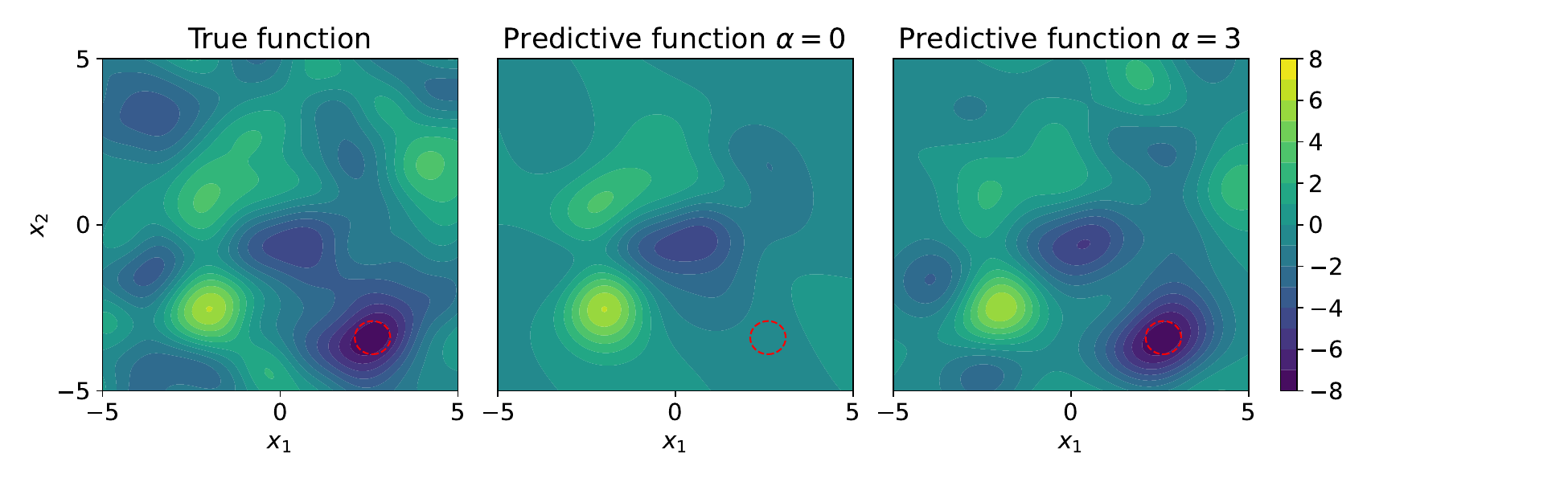}
    \centering{\quad (a) cross-section at $x_3 = 2.5$}
    \end{minipage}
    \begin{minipage}[b]{\linewidth}
    \includegraphics[width = \linewidth]{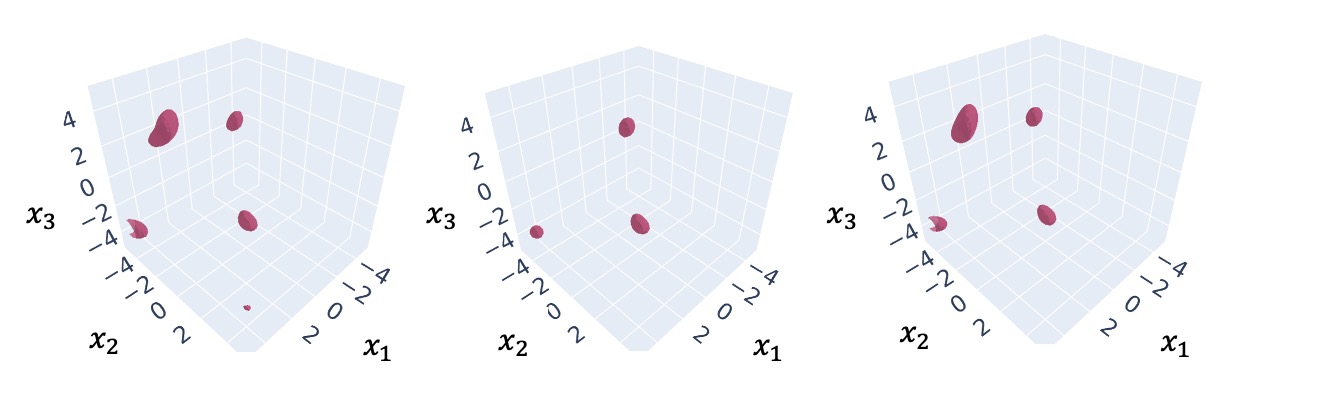}
    \centering{\quad (b) level set $f=-6$}
    \end{minipage}
    \caption{Results for a typical 3D RBF function after 396 sequential samples. (Left column) True function, (middle column) predicted function with $\alpha=0$ and (right column) predicted function with $\alpha=3$, visualized (a) on a cross-section at $x_3=2.5$ and (b) in terms of level set $f=-6$. The global minimum of the function around (2.6, -3.4, 2.5) is circled in (a). }
    \label{fig:3d_case}
\end{figure}

We further test the performance of $acq_{GLW}$ on 3D RBF functions, with one example of these functions shown in figure \ref{fig:3d_func} visualized through level sets of the function. It is clear that the multi-modal feature is still present in the 3D case, which needs to be captured in sampling to resolve the tail of the response PDF. For these 3D functions, our computations start from 8 initial samples followed by 392 sequential samples.

Like in 2D cases, the variation of $t$ with $\alpha=0$ does not affect the performance of $acq_{GLW}$ (figure \ref{fig:3d_RBF}(a)). The major improvement in performance is achieved by increasing $\alpha$ to about 3, for which the error $\epsilon$ after 392 sequential samples is about one order of magnitude smaller than that from $\alpha=0$, corresponding to $acq_{LW}$ (figure \ref{fig:3d_RBF}(b)). Finally, figure \ref{fig:3d_RBF}(c) shows that $\alpha\approx 3$ and $t\approx 1$ still provides the globally optimal results for these 3D functions. 

The mechanism underlying the improved performance with $\alpha=3$ is also similar to the 2D cases. To illustrate this, we consider an example of the RBF function with the global minimum of value $-7.26$ at $\mathbf{x} = (2.6, -3.4, 2.5)$. The predicted response functions with $\alpha=0$ and $\alpha=3$ after 392 samples are shown in figure \ref{fig:3d_case}, visualized respectively on the cross-section at $x_3=2.5$ in (a) and in terms of level set of value $-6$ in (b). It is clear that the global minimum of the function is only captured with $\alpha=3$ and completely missed with $\alpha=0$.

\subsection{Rare-event statistics of ship motion in a random sea}

We finally consider an application of our method to an engineering problem of estimating the rare-event statistics of ship roll motion in a random sea. To simulate the ship roll response in waves, we use a phenomenological nonlinear roll equation that is widely used in marine engineering \cite{umeda2004nonlinear, spyrou2008problems, gong2021full, gong2022sequential, gong2022efficient}
\begin{equation}
     \ddot{\xi}+\alpha_1 \dot{\xi}+\alpha_2 \dot{\xi}|\dot{\xi}| +(\beta_1+\epsilon_1 \sin (\gamma) \eta(t) )\xi+\beta_2 \xi^3 = \epsilon_2 \cos (\gamma) \eta(t), 
    \label{roll}
\end{equation}
where $\xi(t)$ is the time series of roll motion excited by waves with elevation $\eta(t)$, $\gamma$ is the angle between ship heading direction and the wave crest. The empirical coefficients in \eqref{roll} are set as $\alpha_1=0.1$, $\alpha_2=0.1$, $\beta_1=1$, $\beta_2=0.1$, $\epsilon_1=1$, $\epsilon_2=1$. 

\begin{figure}
    \centering
    \includegraphics[width=\linewidth]{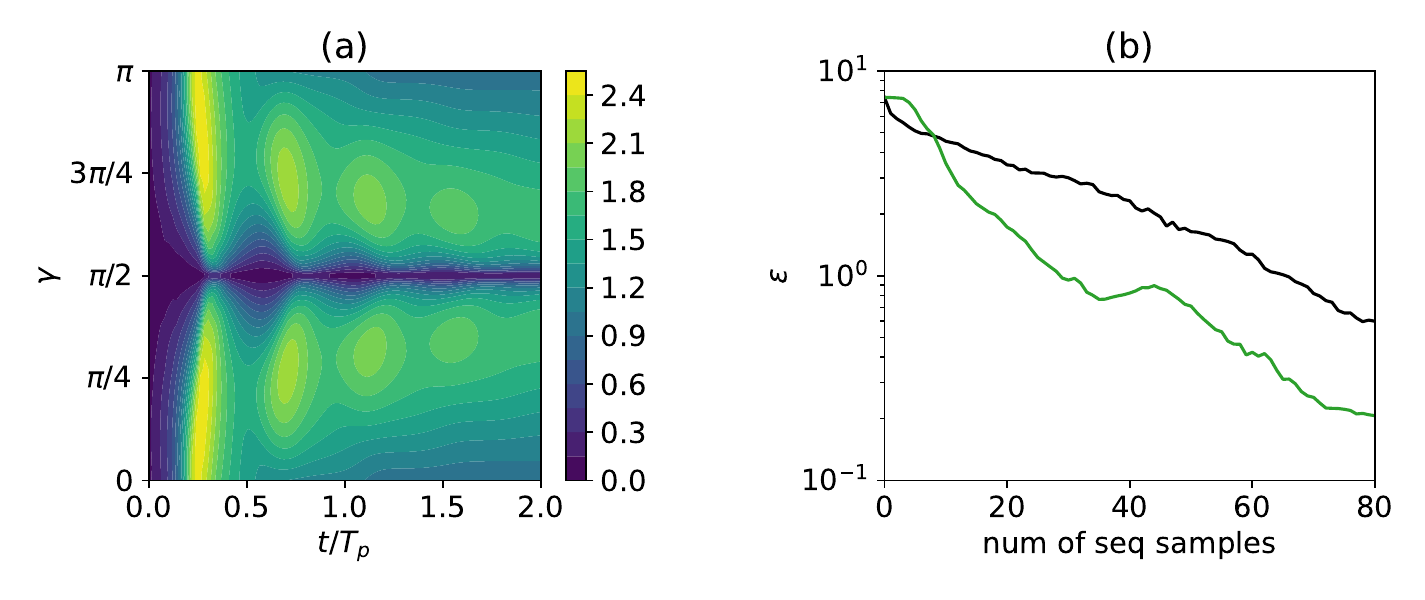}
    \caption{(a) Contour plot of the true response function calculated by \eqref{roll} and (b) results with  $\alpha=3$ (\greenline) and $\alpha=0$ (\blackline) for comparison, both with $t=1$.}
    \label{fig:ship}
\end{figure}

Since large ship motions are usually excited by wave groups at sea, we consider $\eta(t)$ modeled by wave groups with Gaussian envelop 
\begin{equation}
\eta(t) = \exp(-\frac{1}{2}(\frac{t-5T}{2T})^2) \sin(\frac{2\pi}{T}t),
\end{equation}
with $T$ the period of each individual wave in the group. In a random sea, we further consider two independent random parameters $(T, \gamma)$ as the input space, satisfying the distribution of $T \sim \mathcal{N}(T_p, (T_p/4)^2)$ with $T_p=15s$ and $\gamma \sim \mathcal{N}(\gamma_p, (\gamma_p/4)^2)$ with $\gamma_p = \pi/2$.  Our quantity of interest is the maximum roll over the time window $[0, 10T]$:
\begin{equation} 
    f(\mathbf{x}) = \mathrm{max}_{t \in [0, \, 10T]} \; |\xi(t; \mathbf{x})|, 
\label{roll:f}
\end{equation}
with the response function plotted in figure \ref{fig:ship}(a) clearly showing the multi-modal feature.

\begin{figure}
    \centering
    \begin{minipage}[b]{0.48\linewidth}
    \includegraphics[width = \linewidth]{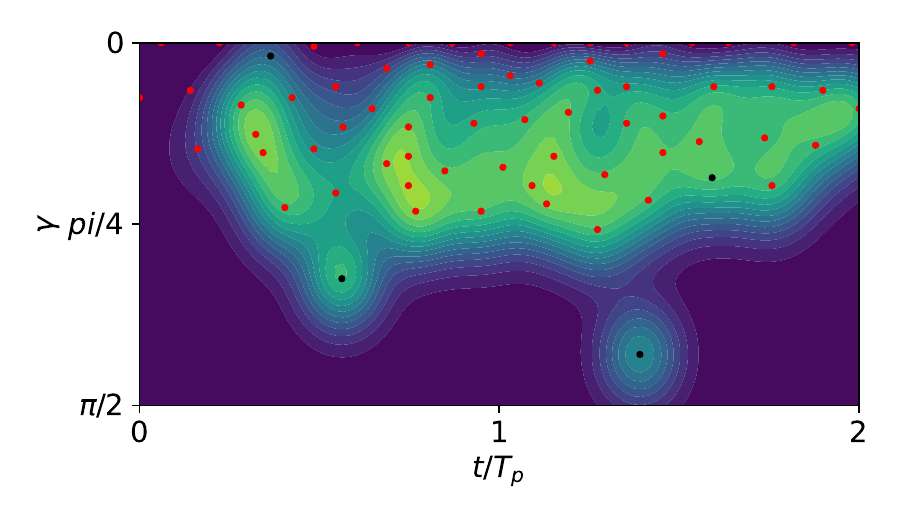}
    \centering{\quad (a)}
    \end{minipage}
    \begin{minipage}[b]{0.48\linewidth}
    \includegraphics[width = \linewidth]{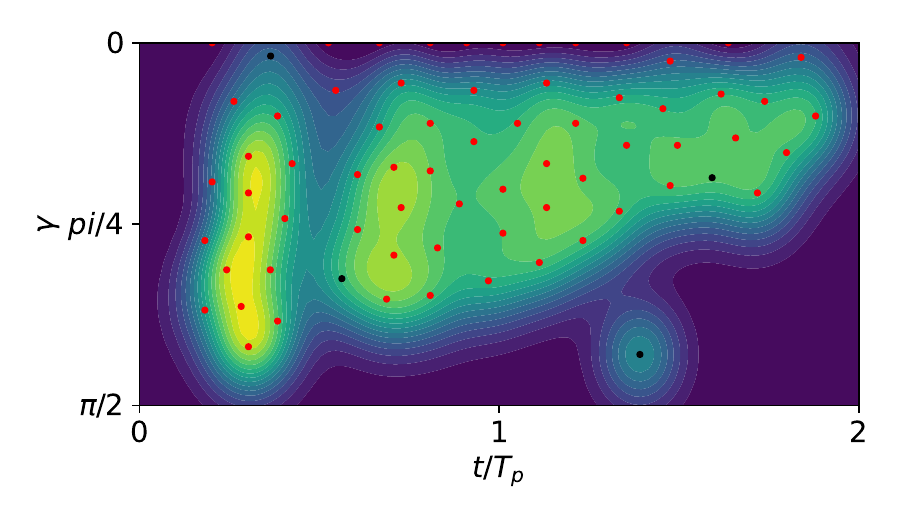}
    \centering{\quad (b)}
    \end{minipage}
    \caption{Predicted response functions and sequential sampling locations (\tikzcircle{1pt, red}) with $t=1$ and (a) $\alpha=$ 0, (b) $\alpha=3$ in the ship motion example, starting from the same initial samples (\tikzcircle{1pt, black}).}
    \label{fig:ship_samples}
\end{figure}

In computation, we only consider half of the input space due to symmetry of \eqref{roll} with $\gamma=\pi/2$. The results from sequential sampling with $acq_{GLW}$ are shown in figure \ref{fig:ship}(b), comparing the cases with $\alpha=3$ and $\alpha=0$. For such a multi-modal response function, it is clear that the result from $\alpha=3$ (empirically determined as optimal in \S4.2) is much better than that from $\alpha=0$, with the error $\epsilon$ from the former half an order of magnitude smaller than the latter in majority of the sampling process. The sampling location plotted in figure \eqref{fig:ship_samples} further demonstrates the effectiveness of $\alpha=3$ to explore rare-event regions in the full input space.

\section{Conclusion and Discussion}
In this work, we develop a new acquisition function $acq_{GLW}$ in sequential sampling to efficiently quantify the rare-event statistics in the response of an ItR system. Our new acquisition takes a generalized form of the existing likelihood-weighed acquisition $acq_{LW}$ \cite{sapsis2020output, sapsis2022optimal} and contains two additional parameters $\alpha$ and $t$. By varying $\alpha$ and $t$, $acq_{GLW}$ is able to (i) place different level of emphasis on rare-event regions in sampling, and (ii) remedy the situation when the predicted ItR function has a large discrepancy from the true function. We demonstrate the advantage of $acq_{GLW}$ over $acq_{LW}$ in a number of test cases with empirically optimal values of $\alpha$ and $t$ identified. 

The test cases include two cases with relatively simple response functions that were previously constructed in \cite{blanchard2020output, sapsis2022optimal} and \cite{pickering2022discovering}, a case with large numbers of complex multi-modal response functions generated from the RBF and Matern kernels, and an engineering case to quantify the rare-event ship roll statistics in a random sea. It is suggested in these cases that if the response function is relatively simple, using $\alpha=1$ and $t\in[1.2,1.6]$ in $acq_{GLW}$ produces consistently better results than that from $acq_{LW}$, due to the more appropriate emphasis on the known rare-event region. If the response function is complex with multi-modal structures, using $\alpha \approx 3$ and $t \approx 1$ is critical since it allows more exploration in sampling to identify multiple rare-event regions in the input space. While these rule-of-thumb values are helpful in applying $acq_{GLW}$, it may be more desirable to develop other advanced methods to automatically determine optimal values of $\alpha$ and $t$ for any given case. One idea here is to apply reinforcement learning to train a policy function $(\alpha, t) = \pi(\mathcal{D})$ so that the optimal $\alpha$ and $t$ can be sampled given the existing dataset (which also contains information on the feature of the response function). In such a manner, $\alpha$ and $t$ can also vary in the sampling process for a given case, achieving even better performance in quantification of the rare-event statistics. We leave this task to our future work. 

We finally point out that the idea of using LW factor in acquisition functions has been widely extended to applications other than rare-event statistics quantification. These applications, as mentioned in \S 1, include rare-event forecasting \cite{pickering2022discovering, rudy2023output}, Bayesian optimization \cite{blanchard2021bayesian}, robot path planning \cite{blanchard2020informative}, multi-arm bandit \cite{yang2022output} and has been adapted to multi-fidelity context \cite{gong2022multi}. We expect that the generalization developed in this paper should apply equally well to these cases, upon more tests to be done for confirmation.

\section*{ACKNOWLEDGEMENT}
We thank the support from the Office of Naval Research grant N00014-23-1-2427. 

\bibliographystyle{unsrt}
\bibliography{reference.bib}

\appendix

\section{Gaussian process regression}
\label{app:gpr}
In this section, we briefly introduce the Gaussian process regression (GPR) \cite{rasmussen2003gaussian}, which is a probabilistic machine learning approach. Consider the task of inferring $f$ from dataset $\mathcal{D}=\{\mathbf{X}, \mathbf{y}\}$. In GPR, a prior, representing our beliefs over all possible functions we expect to observe, is placed on $f$ as a Gaussian process $f(\mathbf{x}) \sim \mathcal{GP}(0, k(\mathbf{x},\mathbf{x}'))$ with zero mean and covariance function $k$. Following the Bayes' theorem, the posterior prediction for $f$ given the dataset $\mathcal{D}$ can be derived to be another Gaussian:  
\begin{equation}
    f(\mathbf{x})|\mathcal{D} \sim \mathcal{GP}\big(\mathbb{E}(f(\mathbf{x})|\mathcal{D}), {\rm{cov}}(f(\mathbf{x}), f(\mathbf{x}')|\mathcal{D})\big),
\label{sgp1}
\end{equation}
with mean and covariance respectively:
\begin{align}
     \mathbb{E}(f(\mathbf{x})|\mathcal{D}) &=  k(\mathbf{x}, \mathbf{X}){\rm{K}}(\mathbf{X},\mathbf{X})^{-1} \mathbf{y}, 
\label{sgp2} \\
     {\rm{cov}}(f(\mathbf{x}), f(\mathbf{x}')|\mathcal{D}) &=k(\mathbf{x},\mathbf{x}')  - k(\mathbf{x},\mathbf{X})  {\rm{K}}(\mathbf{X},\mathbf{X})^{-1} k(\mathbf{X}, \mathbf{x}'), 
\label{sgp3}
\end{align}
where matrix element ${\rm{K}}(\mathbf{X},\mathbf{X})_{ij}=k(\mathbf{x}^i,\mathbf{x}^j)$. 

For covariance functions $k$, we use either radial-basis-function (RBF) kernel or Matern kernel in this paper, respectively defined as 
\begin{equation}
    k (\mathbf{x},\mathbf{x}') = \tau^2 {\rm{exp}}(-\frac{1}{2} \mathrm{dist}^2(\mathbf{x}, \mathbf{x}') ), 
\label{RBF}
\end{equation}
and 
\begin{equation}
    k (\mathbf{x},\mathbf{x}') = \frac{\tau^2}{\Gamma(\nu) 2^{\nu-1}} (\sqrt{2\nu} \; \mathrm{dist}(\mathbf{x}, \mathbf{x}'))^\nu K_\nu\big(\sqrt{2\nu} \; \mathrm{dist}(\mathbf{x}, \mathbf{x}')\big).
\label{Matern}
\end{equation}
The dist function in \eqref{RBF} and  \eqref{Matern} is computed by:
\begin{equation}
    \mathrm{dist}(\mathbf{x}, \mathbf{x}') = ((\mathbf{x}- \mathbf{x}')^T \Lambda^{-1}(\mathbf{x}- \mathbf{x}'))^{\frac{1}{2}},
\end{equation}
where $\tau$ and diagonal matrix $\Lambda$ are hyperparameters representing the characteristic amplitude and length scales. For Matern kernel, $K_{\nu}(\cdot)$ is a modified Bessel function, and $\Gamma(\cdot)$ is the gamma function. $\nu$ is a pre-defined parameter controlling the continuity of the realizations where a smaller value indicates a less smooth function. As $\nu \rightarrow \infty$, the Matern kernel becomes equivalent to the RBF kernel (infinitely differentiable) while $\nu=1.5$ and 2.5 respectively indicate once and twice differentiable functions.  

The hyperparameters $\tau$ and $\Lambda$ in these kernels are determined by maximizing the likelihood $p(\mathbf{y})$.

\section{Derivation of \eqref{approx_L}}
\label{app:approx_L}
The derivation of \eqref{approx_L} is built on the Theorem 2 in \cite{mohamad2018sequential}, restated here with slight change of notations in the context of the current paper.

\textbf{Theorem}: \textit{Let $p_{f^{\pm}|\mathcal{D}, \hat{f}(\tilde{\mathbf{x}})}(f)$ be PDF bounds generated by upper and lower bounds of GPR $f|\mathcal{D}, \hat{f}(\tilde{\mathbf{x}})$. Assume $\mathrm{std}(\mathbf{x}|\mathcal{D}, \hat{f}(\tilde{\mathbf{x}}))$ is sufficiently small (thus $p_{f^{\pm}}(f)$ are close enough). The integration of log difference between $p_{f^{\pm}}(f)$ in \eqref{e_L} is then given by} 
\begin{align}
    \epsilon_{L}(\tilde{\mathbf{x}}) & = \int |\log p_{f^{+}|\mathcal{D}, \hat{f}(\tilde{\mathbf{x}})}(s_{}) - \log p_{f^{-}|\mathcal{D}, \hat{f}(\tilde{\mathbf{x}})}(s_{})| \mathrm{d} s_{}
\nonumber \\ 
    & \approx \int \Big| \frac{\frac{\mathrm{d}}{\mathrm{d}s} \int \mathrm{std}(\mathbf{x}|\mathcal{D}, \hat{f}(\tilde{\mathbf{x}})) p_{\mathbf{x}}(\mathbf{x}) \delta(s- \hat{f}(\mathbf{x})) \mathrm{d} \mathbf{x} }{p_{\hat{f}}(s)} \Big| \mathrm{d}s.
\label{e_L_1}
\end{align}

Let $g(s,\tilde{\mathbf{x}}) \equiv \int \mathrm{std}(\mathbf{x}|\mathcal{D}, \hat{f}(\tilde{\mathbf{x}})) 
 p_{\mathbf{x}}(\mathbf{x})   \delta(s- \hat{f}(\mathbf{x})) \mathrm{d} \mathbf{x}$ and denote $\partial g(s, \tilde{\mathbf{x}})/ \partial s$ as $g'(s, \tilde{\mathbf{x}})$, $\epsilon_{L}$ in \eqref{e_L_1} can be further computed as
\begin{align}
    \epsilon_{L}(\tilde{\mathbf{x}}) & \approx \int_{g'(s, \tilde{\mathbf{x}})>0}  \frac{g'(s, \tilde{\mathbf{x}}) }{p_{\hat{f}}(s)}  \mathrm{d}s - \int_{g'(s, \tilde{\mathbf{x}})<0}  \frac{g'(s, \tilde{\mathbf{x}})}{p_{\hat{f}}(s)}  \mathrm{d}s 
\nonumber \\
     & \approx B + \int_{g'(s, \tilde{\mathbf{x}})>0}  \frac{p'_{\hat{f}}(s)}{p^2_{\hat{f}}(s)} g(s,\tilde{\mathbf{x}}) \mathrm{d}s - \int_{g'(s, \tilde{\mathbf{x}})<0}  \frac{p'_{\hat{f}}(s)}{p^2_{\hat{f}}(s)} g(s,\tilde{\mathbf{x}})  \mathrm{d}s,
\label{appe_L_2}
\end{align}
where we have used integration by parts, with all boundary terms collected in $B$. We note that since there are only finite number of boundary terms, $B$ is guaranteed to be bounded. 

Noticing that $g(s,\tilde{\mathbf{x}})>0$ always, we further have from \eqref{appe_L_2}
\begin{align}
    e_{L} (\tilde{\mathbf{x}}) & \leq B + \int \frac{|p'_{\hat{f}}(s)|}{p^2_{\hat{f}}(s)} g(s,\tilde{\mathbf{x}}) \mathrm{d}s, 
\nonumber \\ 
  & \leq    C \int \frac{|p'_{\hat{f}}(s)|}{p^2_{\hat{f}}(s)} \; \int \mathrm{std}(\mathbf{x}|\mathcal{D}, \hat{f}(\tilde{\mathbf{x}})) 
 p_{\mathbf{x}}(\mathbf{x})   \delta(s- \hat{f}(\mathbf{x})) \mathrm{d} \mathbf{x} \; \mathrm{d} s
\nonumber \\
  & = C \iint \frac{|p'_{\hat{f}}(s)|}{p^2_{\hat{f}}(s)} \mathrm{std}(\mathbf{x}|\mathcal{D}, \hat{f}(\tilde{\mathbf{x}})) 
 p_{\mathbf{x}}(\mathbf{x})   \delta(s- \hat{f}(\mathbf{x})) \mathrm{d} s \mathrm{d} \mathbf{x}  
\nonumber \\
    & =  C \int \mathrm{std}(\mathbf{x}|\mathcal{D}, \hat{f}(\tilde{\mathbf{x}})) \frac{p_{\mathbf{x}}(\mathbf{x}) |p'_{\hat{f}}(\hat{f}(\mathbf{x}))| }{p^2_{\hat{f}}(\hat{f}(\mathbf{x}))}  \mathrm{d} \mathbf{x},
\label{e_L_4}
\end{align}
where in the 2nd line we absorb $B$ into another constant $C$ since the two terms in the 1st line are bounded (from above and below). In the third line we have applied the Fubini's theorem and in the fourth line we have integrated out the delta function. \eqref{e_L_4} is exactly \eqref{approx_L} up to a constant.

We note that our derivation outlined above is different from that in \cite{sapsis2022optimal} (in particular their proof of theorem 3.2) which is at least not well understood by the authors.

\section{Acceleration in MCDO regarding the acquisitions}
\label{app:opt}

In MCDO method, we pre-select a large number of candidate samples located at $\mathbf{X}_{mc}\in \mathbb{R}^{n_{mc}*d}$ (usually from space-filling L-H sampling), where $n_{mc}\gg n$ with $n$ the number of samples in the existing dataset $\mathcal{D}$. The optimization problem is then approximated by a discrete optimization
\begin{equation}
    \mathbf{x}^* = \mathrm{argmax}_{\mathbf{x} \in \mathbf{X}_{mc}}  \; acq(\mathbf{x}).
\end{equation}
In the following, we will take \eqref{US-LW} as the acquisition function in presenting the algorithm, but the algorithm applies equally to \eqref{US-GLW} and \eqref{MSE-LW} as discussed in \S3.3. In computing $acq(\mathbf{X}_{mc})$, one needs to evaluate a new GPR with $\hat{f}(\mathbf{X}_{mc}) = \mathbb{E}(\mathbf{X}_{mc}|\mathcal{D})$ and $\mathrm{var}(\mathbf{X}_{mc}|\mathcal{D})$, with the former needed to calculate the function $p_{\hat{f}}(f)$ and its arguments $\hat{f}(\mathbf{x})$. A direct (brute-force) computation following \eqref{sgp2} and \eqref{sgp3} can be conducted as
\small
\begin{subequations}
\begin{align}
    \mathbb{E}(\mathbf{X}_{mc} | \mathcal{D}) & = \mathrm{K}(\mathbf{X}_{mc}, \mathbf{X}) \mathrm{K}(\mathbf{X}, \mathbf{X})^{-1} \mathbf{y}, 
\\
    \mathrm{var}(\mathbf{X}_{mc} | \mathcal{D}) & = \mathrm{diag}\big(\mathrm{K}(\mathbf{X}_{mc}, \mathbf{X}_{mc})\big) - \mathrm{diag}\big(\mathrm{K}(\mathbf{X}_{mc}, \mathbf{X}) \mathrm{K}(\mathbf{X}, \mathbf{X})^{-1}\mathrm{K}(\mathbf{X}, \mathbf{X}_{mc})\big).
\end{align}
\label{full}
\end{subequations}
\normalsize

\sloppy The computational complexity of \eqref{full} consists of three major parts: $(i)$ the Cholesky decomposition of $\mathrm{K}(\mathbf{X}, \mathbf{X}) \in \mathbb{R}^{n^2}$ for computing its inverse $\mathrm{K}(\mathbf{X}, \mathbf{X})^{-1}$, with complexity $O(n^3)$, $(ii)$ obtaining each element in $\mathrm{K}(\mathbf{X}_{mc}, \mathbf{X}) \in \mathbb{R}^{n_{mc} * n} $ with $O(n_{mc} *n)$, $(iii)$ the Cholesky solve of $\mathrm{K}(\mathbf{X}_{mc}, \mathbf{X}) \mathrm{K}(\mathbf{X}, \mathbf{X})^{-1}$ based on results of $(i)$, with complexity $O(n_{mc}* n^2)$. Since $n_{mc} \gg n$, $(iii)$ yields the highest computational complexity among the three procedures, instead of $(i)$ (part of the re-training procedure) which is most computationally intensive for many other applications. In practice, for $n=300$ and $n_{mc}=10^5$ (typical sizes of problems in this paper), only $1\%$ of the total computational time is spent on $(i)$, while $(ii)$ and $(iii)$, on the other hand, contribute approximately equally to the remaining $99\%$ computational time (note that $(ii)$ has a large pre-factor in front of the Big O operator due to the need to compute covariance for each element). Therefore, alleviating the computational cost regarding $(ii)$ and $(iii)$ are most important in developing a fast computational approach. 

Our developed approach leverages the recursive update of GPR used in \cite{blanchard2020output,blanchard2021bayesian,blanchard2020informative, gong2022efficient} with additional techniques of memory-time tradeoff and matrix multiplication strategy. To start, we employ the recursive update of the mean and variance building on that of the last step $f|\mathcal{D}_{n-1}$ (subscript $n-1$ refers to the dataset before the $n$-th sample $\mathbf{x}_{n}$ is added):
\small
\begin{subequations}
\begin{align}
    \mathbb{E}(\mathbf{X}_{mc} | \mathcal{D}_{n-1}, \mathbf{x}_{n}) 
   = &  \; \mathbb{E}(\mathbf{X}_{mc} | \mathcal{D}_{n-1})  + 
    \frac{{\rm{cov}}(\mathbf{X}_{mc},\mathbf{x}_{n}| \mathcal{D}_{n-1})}{{\rm{var}}(\mathbf{x}_n|\mathcal{D}_{n-1})} \big(f(\mathbf{x}_n) - \mathbb{E}(\mathbf{x}_n|\mathcal{D}_{n-1})\big),
\\
    \mathrm{var}(\mathbf{X}_{mc} | \mathcal{D}_{n-1}, \mathbf{x}_{n})  = & \; {\rm{var}}(\mathbf{X}_{mc} | \mathcal{D}_{n-1}) -
    \frac{{\rm{cov}}(\mathbf{X}_{mc},\mathbf{x}_{n} | \mathcal{D}_{n-1})^2}
    {{\rm{var}}(\mathbf{x}_{n} |\mathcal{D}_{n-1})},
\end{align}
\label{recur}
\end{subequations}
\normalsize
with:
\small
\begin{equation}
    \mathrm{cov}(\mathbf{X}_{mc}, \mathbf{x}_{n} | \mathcal{D}_{n-1}) = k(\mathbf{X}_{mc}, \mathbf{x}_{n} ) -\mathrm{K}(\mathbf{X}_{mc}, \mathbf{X}_{n-1}) \mathrm{K}(\mathbf{X}_{n-1}, \mathbf{X}_{n-1})^{-1}k(\mathbf{X}_{n-1}, \mathbf{x}_{n}).
\label{recur_base}
\end{equation}
\normalsize

In \eqref{recur}, we can reuse $\mathbb{E}(\mathbf{X}_{mc} | \mathcal{D}_{n-1})$ and ${\rm{var}}(\mathbf{X}_{mc} | \mathcal{D}_{n-1})$ from last iteration, and the major computational cost lies on \eqref{recur_base}. Here we note that a direct computation of \eqref{recur_base}, as conducted in \cite{blanchard2020output,blanchard2021bayesian,blanchard2020informative} (judged by their uploaded codes in Github\footnote{https://github.com/ablancha/gpsearch}), scales similarly as in $\eqref{full}$. This is because the Cholesky solve step of $(\mathbf{X}_{mc}, \mathbf{X}_{n-1}) \mathrm{K}(\mathbf{X}_{n-1}, \mathbf{X}_{n-1})^{-1}$ costs $O(n_{mc} * (n-1)^2)$ that is similar as $(iii)$ for the brute-force GPR formula \eqref{full}. In order to reduce this part of the computational cost, we can compute \eqref{recur_base} by \textit{parenthesizing in a different way} (see \cite{strang2007computational} for applications in other contexts): first computing $\mathrm{K}(\mathbf{X}_{n-1}, \mathbf{X}_{n-1})^{-1}k(\mathbf{X}_{n-1}, \mathbf{x}_{n}) \in \mathbb{R}^{(n-1)*1}$ and then multiplying the result by $\mathrm{K}(\mathbf{X}_{mc}, \mathbf{X}_{n-1})$. In this way, the original $O(n_{mc}*n^2)$ complexity in $(iii)$ is reduced to $O(n_{mc}*n)$. Regarding $(ii)$, we now do not need to construct $\mathrm{K}(\mathbf{X}_{mc}, \mathbf{X}_{n})$ in \eqref{full}, since only $\mathrm{K}(\mathbf{X}_{mc}, \mathbf{X}_{n-1})$ is involved in \eqref{recur_base} that can be taken from last iteration. The only new construction regards $k(\mathbf{X}_{mc}, \mathbf{x}_n) \in \mathbb{R}^{n_{mc} * 1}$ only takes $O(n_{mc})$ complexity. This is a standard memory-time tradeoff idea where we save $\mathrm{K}(\mathbf{X}_{mc}, \mathbf{X}_{n-1})$ in the memory, with the advantage of greatly reducing the computational requirement. 

\begin{figure}
    \centering
    \includegraphics[width=0.6\linewidth]{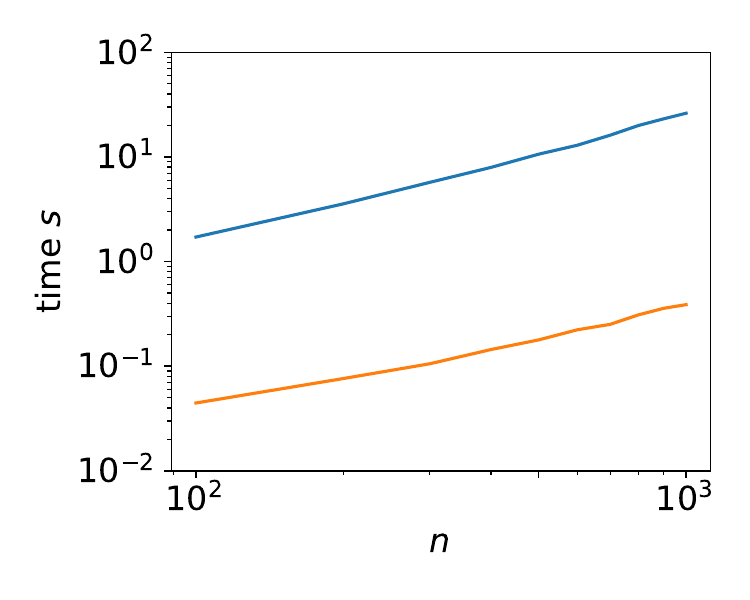}
    \caption{Computation time for selecting one sequential sample using direct computation \eqref{recur} (\blueline) and the developed approach (\orangeline) for $n_{mc} = 2 * 10^5$ and varying $n$ from 100 to 1000.}
    \label{fig:opt_time}
\end{figure}

To illustrate the superiority of the developed computational method, we show in figure \ref{fig:opt_time} the computation time using the developed approach and direct computation as in \eqref{recur} for $n_{mc}=2*10^5$ and varying $n$ on four cores of Intel Xeon Gold 6154 CPU. It is clear that the developed approach achieves a speedup of one and a half orders of magnitude. 

\section{Results for Matern functions}
\label{app:Matern}
\begin{figure}[H]
    \centering
    \begin{minipage}[b]{0.48\linewidth}
    \includegraphics[width = \linewidth]{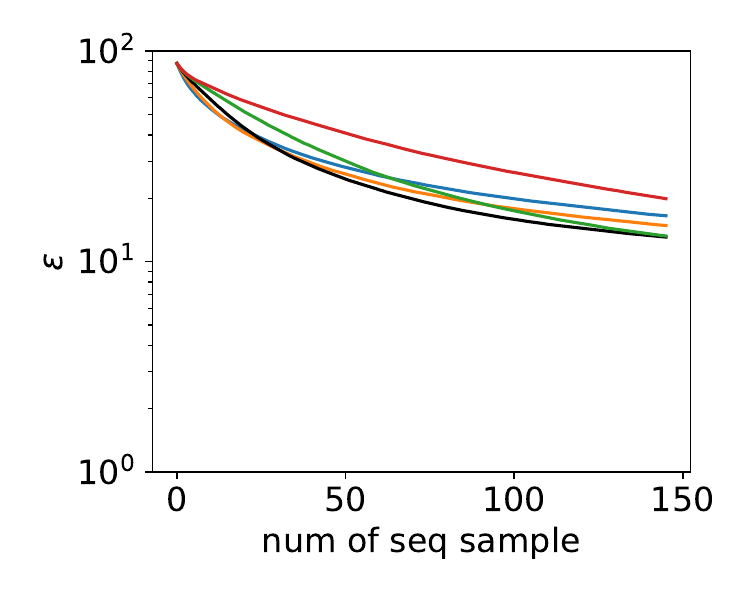}
    \centering{\quad (a)}
    \end{minipage}
    \begin{minipage}[b]{0.48\linewidth}
    \includegraphics[width = \linewidth]{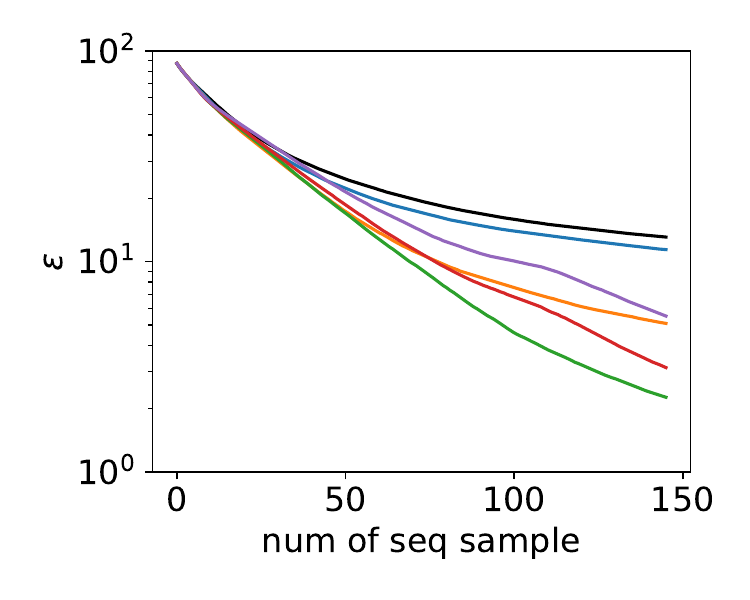}
    \centering{\quad (b)}
    \end{minipage}
   \begin{minipage}[b]{0.6\linewidth}
    \includegraphics[width = \linewidth]{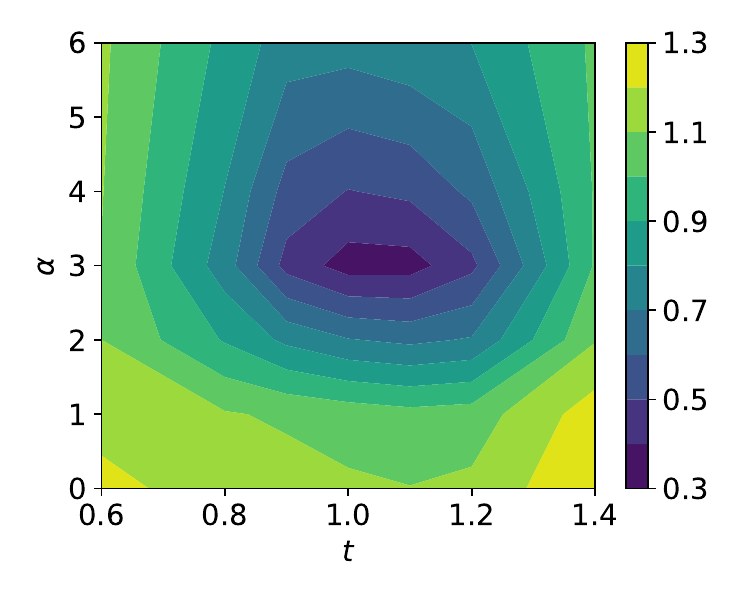}
    \centering{\quad (c)}
    \end{minipage}
    \caption{Results for two-dimensional Matern functions. Error $\epsilon$ as function of number of samples for (a) $\alpha=0$ and varying $t=$ 0.6 (\blueline), 0.8(\orangeline), 1(\blackline), 1.2(\greenline), 1.4(\redline), (b) $t=1$ and varying $\alpha=$ 0(\blackline),  1 (\blueline), 2(\orangeline), 3(\greenline), 4(\redline), 6(\purpleline); (c) contour plot of $\log_{10} \epsilon$ at 146 sequential samples for varying $t$ and $\alpha$.}
    \label{fig:2d_Matern}
\end{figure}

\begin{figure}
    \centering
    \includegraphics[width=\linewidth]{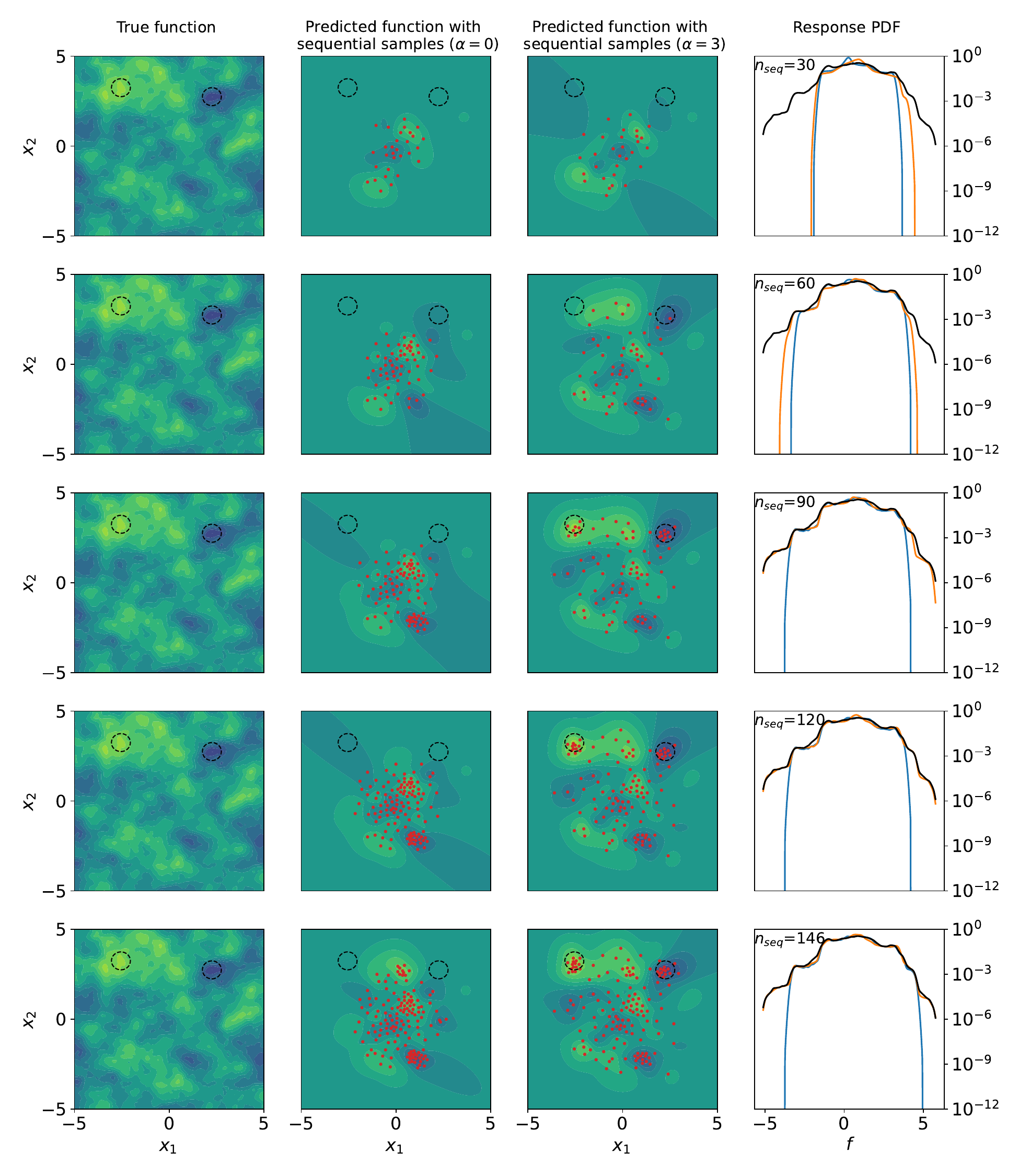}
    \caption{First column: true response Matern function as a reference; second column: sequential samples (\tikzcircle{2pt, Red}) with $\alpha=0$ on the predicted response function; third column: sequential samples (\tikzcircle{2pt, Red}) with $\alpha=3$ on the predicted response function; fourth column: predicted PDF $p_{\hat{f}}(f)$ with $\alpha=0$ (\blueline) and $\alpha=3$ (\orangeline) compared with the true PDF $p_f(f)$ (\blackline). The top-to-bottom rows correspond to situations with number of sequential samples $n_{seq} = [30, 60, 90, 120, 146]$. The black circles shown in columns 1-3 mark the rare-event regions around $(-2.6, 3.2)$ and $(2.3, 2.8)$ that are missed by sequential samples with $\alpha=0$ but captured with $\alpha=3$. } 
    \label{fig:evolution_matern}
\end{figure}

\begin{figure}
    \centering
    \begin{minipage}[b]{0.48\linewidth}
    \includegraphics[width = \linewidth]{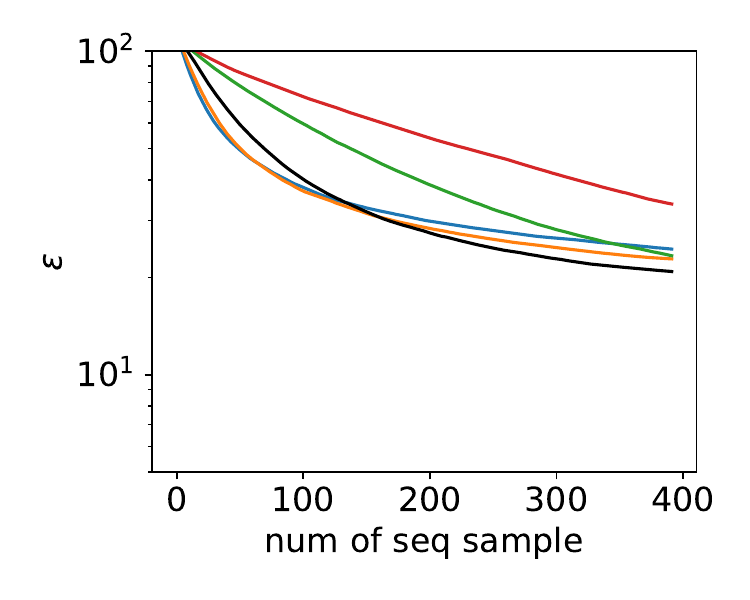}
    \centering{\quad \quad (a) }
    \end{minipage}
    \begin{minipage}[b]{0.48\linewidth}
    \includegraphics[width = \linewidth]{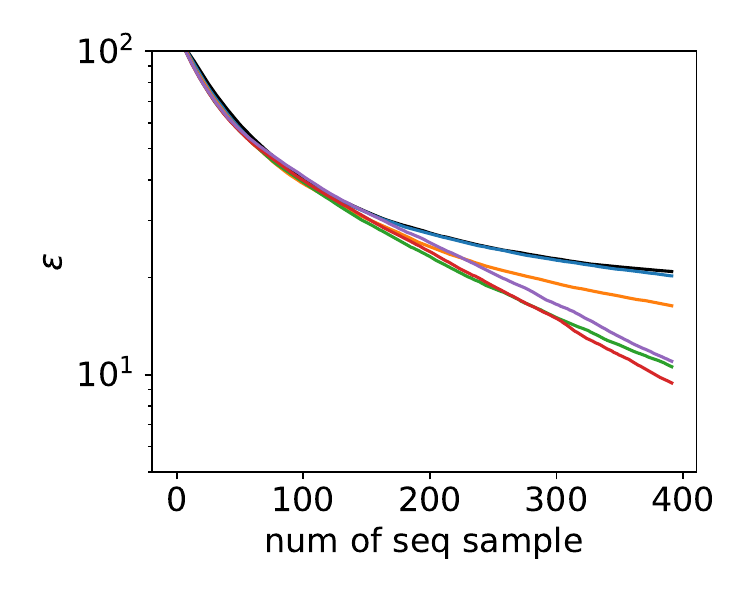}
    \centering{\quad \quad  (b) }
    \end{minipage}
    \begin{minipage}[b]{0.6\linewidth}
    \includegraphics[width = \linewidth]{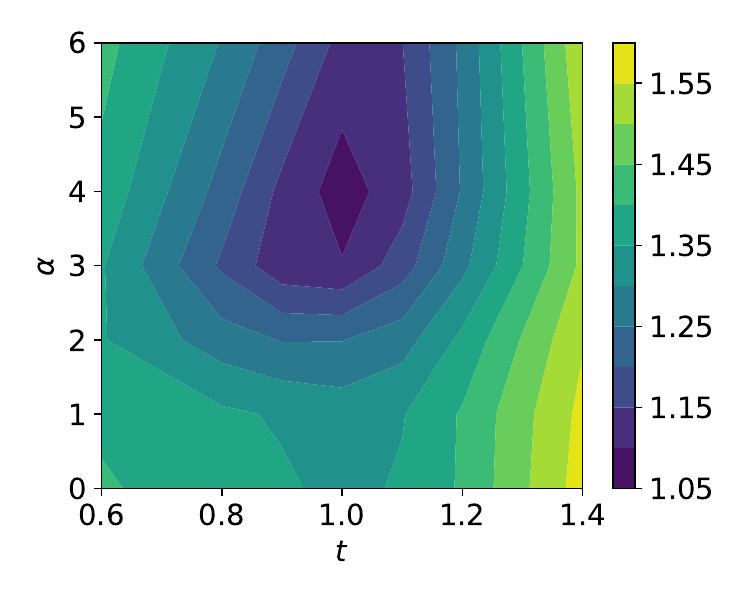}
    \centering{(c)}
    \end{minipage}
    \caption{Results for three-dimensional Matern functions. Error $\epsilon$ as function of number of samples for (a) $\alpha=0$ and varying $t=$ 0.6 (\blueline), 0.8(\orangeline), 1(\blackline), 1.2(\greenline), 1.4(\redline), (b) $t=1$ and varying $\alpha=$ 0(\blackline),  1 (\blueline), 2(\orangeline), 3(\greenline), 4(\redline), 6(\purpleline); (c) contour plot of $\log_{10}\epsilon$ at 392 sequential samples for varying $t$ and $\alpha$.
    }
    \label{fig:3d_Matern} 
\end{figure}

In this section, we collect the results for two and three-dimensional Matern functions mentioned in \S4.2. Figures \ref{fig:2d_Matern}, \ref{fig:evolution_matern}, and \ref{fig:3d_Matern} respectively correspond to figures \ref{fig:2d_RBF}, \ref{fig:evolution_RBF}, and \ref{fig:3d_RBF}, but with the response functions (and GPR) generated by the Matern kernel \eqref{Matern}. Conclusions from these cases with Matern functions are very similar to what we reach in \S4.2 for the RBF functions. The only comment needed is that for the 3D Matern functions, the global optimal of $\alpha$ and $t$ is achieved at $\alpha \approx 4$ and $t \approx 1$ instead of $\alpha \approx 3$ and $t \approx 1$ as in the RBF cases. However, the latter still provides a near-optimal performance for the Matern cases.  

\end{document}

%% file: color.tex
\usepackage{color}
\usepackage[dvipsnames]{xcolor}
\usepackage{tikz}

\DeclareRobustCommand{\blackline}{\raisebox{2pt}{\tikz{\draw[-,black!40!Black,solid,line width = 0.9pt](0,0) -- (5mm,0);}}}
\DeclareRobustCommand{\blueline}{\raisebox{2pt}{\tikz{\draw[RoyalBlue,solid,line width = 0.9pt](0,0) -- (5mm,0);}}}
\DeclareRobustCommand{\orangeline}{\raisebox{2pt}{\tikz{\draw[BurntOrange,solid,line width = 0.9pt](0,0) -- (5mm,0);}}}
\DeclareRobustCommand{\greenline}{\raisebox{2pt}{\tikz{\draw[Green,solid,line width = 0.9pt](0,0) -- (5mm,0);}}}
\DeclareRobustCommand{\redline}{\raisebox{2pt}{\tikz{\draw[Red,solid,line width = 0.9pt](0,0) -- (5mm,0);}}}
\DeclareRobustCommand{\purpleline}{\raisebox{2pt}{\tikz{\draw[Purple,solid,line width = 0.9pt](0,0) -- (5mm,0);}}}

\DeclareRobustCommand{\bluedashedline}{\raisebox{2pt}{\tikz{\draw[RoyalBlue,dashed,line width = 0.9pt](0,0) -- (5mm,0);}}}
\DeclareRobustCommand{\orangedashedline}{\raisebox{2pt}{\tikz{\draw[BurntOrange,dashed,line width = 0.9pt](0,0) -- (5mm,0);}}}
\DeclareRobustCommand{\greendashedline}{\raisebox{2pt}{\tikz{\draw[Green,dashed,line width = 0.9pt](0,0) -- (5mm,0);}}}

\DeclareRobustCommand{\bluedashdotline}{\raisebox{2pt}{\tikz{\draw[RoyalBlue,densely dash dot,line width = 0.9pt](0,0) -- (5mm,0);}}}
\DeclareRobustCommand{\orangedashdotline}{\raisebox{2pt}{\tikz{\draw[BurntOrange,densely dash dot,line width = 0.9pt](0,0) -- (5mm,0);}}}
\DeclareRobustCommand{\greendashdotline}{\raisebox{2pt}{\tikz{\draw[Green,densely dash dot,line width = 0.9pt](0,0) -- (5mm,0);}}}
\DeclareRobustCommand{\reddashdotline}{\raisebox{2pt}{\tikz{\draw[Red,densely dash dot,line width = 0.9pt](0,0) -- (5mm,0);}}}

\DeclareRobustCommand{\tikzcircle}[2][red,fill=red]{\tikz[baseline=-0.5ex]\draw[#1,radius=#2] (0,0) circle ;}